\documentclass[final]{cvpr}

\usepackage[utf8]{inputenc} %
\usepackage[T1]{fontenc}    %
\usepackage{url}            %
\usepackage{booktabs}       %
\usepackage{amsfonts}       %
\usepackage{nicefrac}       %
\usepackage{microtype}      %

\usepackage{times}
\usepackage{epsfig}
\usepackage{graphicx}
\usepackage{comment}
\usepackage{multirow}
\usepackage{lipsum,pdflscape}
\usepackage{rotating}
\usepackage{subcaption}
\usepackage{footnote}
\usepackage{amsmath,amssymb} %
\usepackage{color}
\usepackage[ruled,linesnumbered, vlined]{algorithm2e}
\DeclareGraphicsExtensions{.pdf}
\usepackage{enumitem}  %
\usepackage{cite}  %
\usepackage{threeparttable}

\usepackage{soul}

\usepackage{xcolor}

\usepackage{float}
\restylefloat{table}

\usepackage{pdfrender}

\definecolor{forestgreen}{rgb}{0.13, 0.55, 0.13}

\usepackage[pagebackref=true,breaklinks=true,colorlinks,bookmarks=false]{hyperref}

\def\confYear{CVPR 2021}
\begin{document}

\title{NetAdaptV2: Efficient Neural Architecture Search with \\ Fast Super-Network Training and Architecture Optimization\vspace{-1em}}

\author{Tien-Ju Yang, Yi-Lun Liao, Vivienne Sze\\
Massachusetts Institute of Technology\\
{\tt\small \{tjy,ylliao,sze\}@mit.edu}}

\maketitle

\begin{abstract}

Neural architecture search (NAS) typically consists of three main steps: training a super-network, training and evaluating sampled deep neural networks (DNNs), and training the discovered DNN. Most of the existing efforts speed up some steps at the cost of a significant slowdown of other steps or sacrificing the support of non-differentiable search metrics. The unbalanced reduction in the time spent per step limits the total search time reduction, and the inability to support non-differentiable search metrics limits the performance of discovered DNNs.

In this paper, we present NetAdaptV2 with three innovations to better balance the time spent for each step while supporting non-differentiable search metrics. First, we propose channel-level bypass connections that merge network depth and layer width into a single search dimension to reduce the time for training and evaluating sampled DNNs. Second, ordered dropout is proposed to train multiple DNNs in a single forward-backward pass to decrease the time for training a super-network. Third, we propose the multi-layer coordinate descent optimizer that considers the interplay of multiple layers in each iteration of optimization to improve the performance of discovered DNNs while supporting non-differentiable search metrics. With these innovations, NetAdaptV2 reduces the total search time by up to $5.8\times$ on ImageNet and $2.4\times$ on NYU Depth V2, respectively, and discovers DNNs with better accuracy-latency/accuracy-MAC trade-offs than state-of-the-art NAS works. Moreover, the discovered DNN outperforms NAS-discovered MobileNetV3 by 1.8\% higher top-1 accuracy with the same latency.\footnote{The project website: \url{http://netadapt.mit.edu}.}

\end{abstract}

\begin{figure}[t]
\begin{center}
   \includegraphics[width=1.0\linewidth]{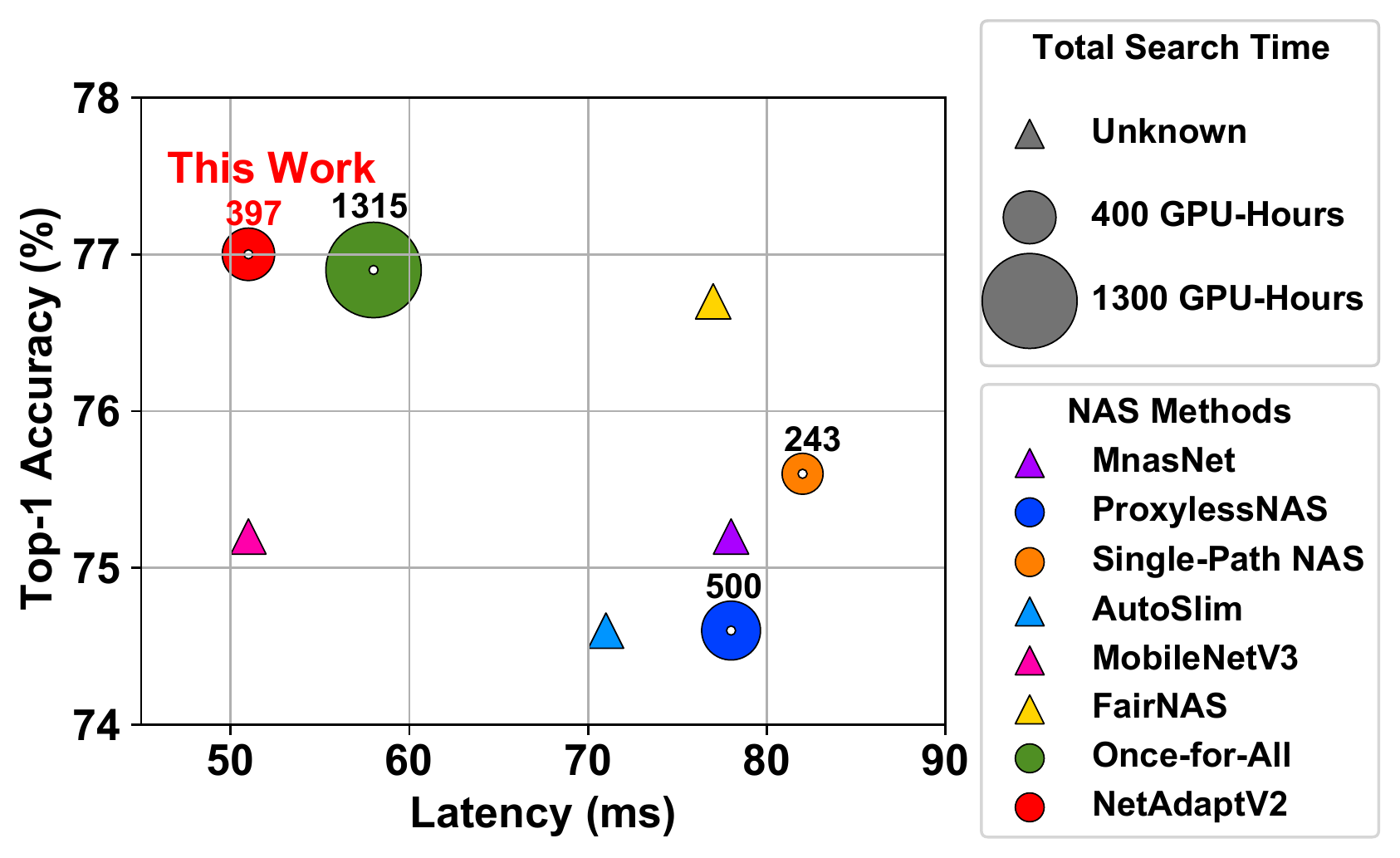}
\end{center}
   \vspace{-4mm}
   \caption{The comparison between NetAdaptV2 and related works. The number above a marker is the corresponding total search time measured on NVIDIA V100 GPUs.}
\label{fig:nas_comparison}
\end{figure}

\section{Introduction}
\label{sec:introduction}

Neural architecture search (NAS) applies machine learning to automatically discover deep neural networks (DNNs) with better performance (e.g., better accuracy-latency trade-offs) by sampling the search space, which is the union of all discoverable DNNs. The search time is one key metric for NAS algorithms, which accounts for three steps: 1) training a \emph{super-network}, whose weights are shared by all the DNNs in the search space and trained by minimizing the loss across them, 2) training and evaluating sampled DNNs (referred to as \emph{samples}), and 3) training the discovered DNN. Another important metric for NAS is whether it supports non-differentiable search metrics such as hardware metrics (e.g., latency and energy). Incorporating hardware metrics into NAS is the key to improving the performance of the discovered DNNs~\cite{eccv2018-netadapt, Tan2018MnasNetPN, cai2018proxylessnas, Chen2020MnasFPNLL, chamnet}.

There is usually a trade-off between the time spent for the three steps and the support of non-differentiable search metrics. For example, early reinforcement-learning-based NAS methods~\cite{zoph2017nasreinforcement, zoph2018nasnet, Tan2018MnasNetPN} suffer from the long time for training and evaluating samples. Using a super-network~\cite{yu2018slimmable, Yu_2019_ICCV, autoslim_arxiv, cai2020once, yu2020bignas, Bender2018UnderstandingAS, enas, tunas, Guo2020SPOS} solves this problem, but super-network training is typically time-consuming and becomes the new time bottleneck. The gradient-based methods~\cite{gordon2018morphnet, liu2018darts, wu2018fbnet, fbnetv2, cai2018proxylessnas, stamoulis2019singlepath, stamoulis2019singlepathautoml, Mei2020AtomNAS, Xu2020PC-DARTS} reduce the time for training a super-network and training and evaluating samples at the cost of sacrificing the support of non-differentiable search metrics. In summary, many existing works either have an unbalanced reduction in the time spent per step (i.e., optimizing some steps at the cost of a significant increase in the time for other steps), which still leads to a long \emph{total} search time, or are unable to support non-differentiable search metrics, which limits the performance of the discovered DNNs.

In this paper, we propose an efficient NAS algorithm, NetAdaptV2, to significantly reduce the \emph{total} search time by introducing three innovations to \emph{better balance} the reduction in the time spent per step while supporting non-differentiable search metrics:

\textbf{Channel-level bypass connections (mainly reduce the time for training and evaluating samples, Sec.~\ref{subsec:channel_level_bypass_connections})}: Early NAS works only search for DNNs with different numbers of filters (referred to as \emph{layer widths}). To improve the performance of the discovered DNN, more recent works search for DNNs with different numbers of layers (referred to as \emph{network depths}) in addition to different layer widths at the cost of training and evaluating more samples because network depths and layer widths are usually considered independently. In NetAdaptV2, we propose \emph{channel-level bypass connections} to merge network depth and layer width into a single search dimension, which requires only searching for layer width and hence reduces the number of samples.

\textbf{Ordered dropout (mainly reduces the time for training a super-network, Sec.~\ref{subsec:ordered_droput})}: We adopt the idea of super-network to reduce the time for training and evaluating samples. In previous works, \emph{each} DNN in the search space requires one forward-backward pass to train. As a result, training multiple DNNs in the search space requires multiple forward-backward passes, which results in a long training time. To address the problem, we propose \emph{ordered dropout} to jointly train multiple DNNs in a \emph{single} forward-backward pass, which decreases the required number of forward-backward passes for a given number of DNNs and hence the time for training a super-network.

\textbf{Multi-layer coordinate descent optimizer (mainly reduces the time for training and evaluating samples and supports non-differentiable search metrics, Sec.~\ref{subsec:optimizer}):} NetAdaptV1~\cite{eccv2018-netadapt} and MobileNetV3~\cite{Howard_2019_ICCV}, which utilizes NetAdaptV1, have demonstrated the effectiveness of the single-layer coordinate descent (SCD) optimizer~\cite{book2020sze} in discovering high-performance DNN architectures. The SCD optimizer supports both differentiable and non-differentiable search metrics and has only a few interpretable hyper-parameters that need to be tuned, such as the per-iteration resource reduction. However, there are two shortcomings of the SCD optimizer. First, it only considers one layer per optimization iteration. Failing to consider the joint effect of multiple layers may lead to a worse decision and hence sub-optimal performance. Second, the per-iteration resource reduction (e.g., latency reduction) is limited by the layer with the smallest resource consumption (e.g., latency). It may take a large number of iterations to search for a very deep network because the per-iteration resource reduction is relatively small compared with the network resource consumption. To address these shortcomings,  we propose the \emph{multi-layer coordinate descent (MCD) optimizer} that considers multiple layers per optimization iteration to improve performance while reducing search time and preserving the support of non-differentiable search metrics.

Fig.~\ref{fig:nas_comparison} (and Table~\ref{tab:nas_result}) compares NetAdaptV2 with related works. NetAdaptV2 can reduce the search time by up to $5.8\times$ and $2.4\times$ on ImageNet~\cite{imagenet_cvpr09} and NYU Depth V2~\cite{nyudepth} respectively and discover DNNs with better performance than state-of-the-art NAS works. Moreover, compared to NAS-discovered MobileNetV3~\cite{Howard_2019_ICCV}, the discovered DNN has $1.8\%$ higher accuracy with the same latency.

\section{Methodology: NetAdaptV2}

\subsection{Algorithm Overview}

NetAdaptV2 searches for DNNs with different network depths, layer widths, and kernel sizes. The proposed \textbf{channel-level bypass connections (CBCs, Sec.~\ref{subsec:channel_level_bypass_connections})} enables NetAdaptV2 to discover DNNs with different network depths and layer widths by only searching layer widths because different network depths become the natural results of setting the widths of some layers to zero. To search kernel sizes, NetAdaptV2 uses the superkernel method~\cite{stamoulis2019singlepath, stamoulis2019singlepathautoml, yu2020bignas}.

\begin{figure*}[t]
\begin{center}
   \includegraphics[width=1.0\linewidth]{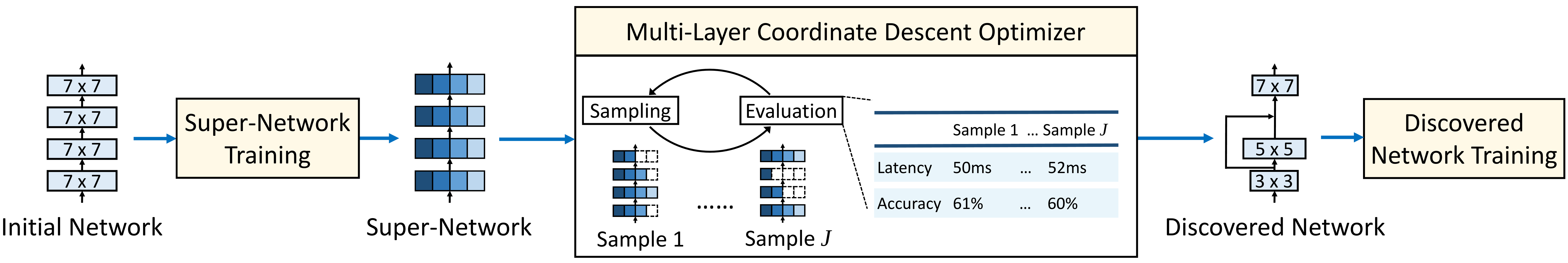}
\end{center}
\vspace{-2mm}
   \caption{The algorithm flow of the proposed NetAdaptV2.}
\label{fig:overview}
\end{figure*}

Fig.~\ref{fig:overview} illustrates the algorithm flow of NetAdaptV2. It takes an initial network and uses its \emph{sub-networks}, which can be obtained by shrinking some layers in the initial network, to construct the search space. In other words, a sample in NetAdaptV2 is a sub-network of the initial network. Because the optimizer needs the accuracy of samples for comparing their performance, the samples need to be trained. NetAdaptV2 adopts the concept of jointly training all sub-networks with shared weights by training a super-network, which has the same architecture as the initial network and contains these shared weights. We use CBCs, the proposed \textbf{ordered dropout (Sec.~\ref{subsec:ordered_droput})}, and superkernel~\cite{stamoulis2019singlepath, stamoulis2019singlepathautoml, yu2020bignas} to efficiently train the super-network that contains sub-networks with different layer widths, network depths, and kernel sizes. After training the super-network, the proposed \textbf{multi-layer coordinate descent optimizer (Sec.~\ref{subsec:optimizer})} is used to discover the architectures of DNNs with optimal performance. The optimizer iteratively samples the search space to generate a bunch of samples and determines the next set of samples based on the performance of the current ones. This process continues until the given stop criteria are met (e.g., the latency is smaller than 30ms), and the discovered DNN is then trained until convergence. Because of the trained super-network, the accuracy of samples can be directly evaluated by using the shared weights without any further training.

\subsection{Channel-Level Bypass Connections}
\label{subsec:channel_level_bypass_connections}

Previous NAS algorithms generally treat network depth and layer width as two different search dimensions. The reason is evident in the following example. If we remove a filter from a layer, we reduce the number of output channels by one. As a result, if we remove all the filters, there are no output channels for the next layer, which breaks the DNN into two disconnected parts. Hence, reducing layer widths typically cannot be used to reduce network depths. To address this, we need an approach that keeps the network connectivity while removing filters; this is achieved by our proposed channel-level bypass connections (CBCs).

\begin{figure*}[t]
    \centering
    \subfloat[Case 1: Same number of input channels and initial filters. \\ ($C = T = 4$)]
    {
        \includegraphics[width=0.225\linewidth]{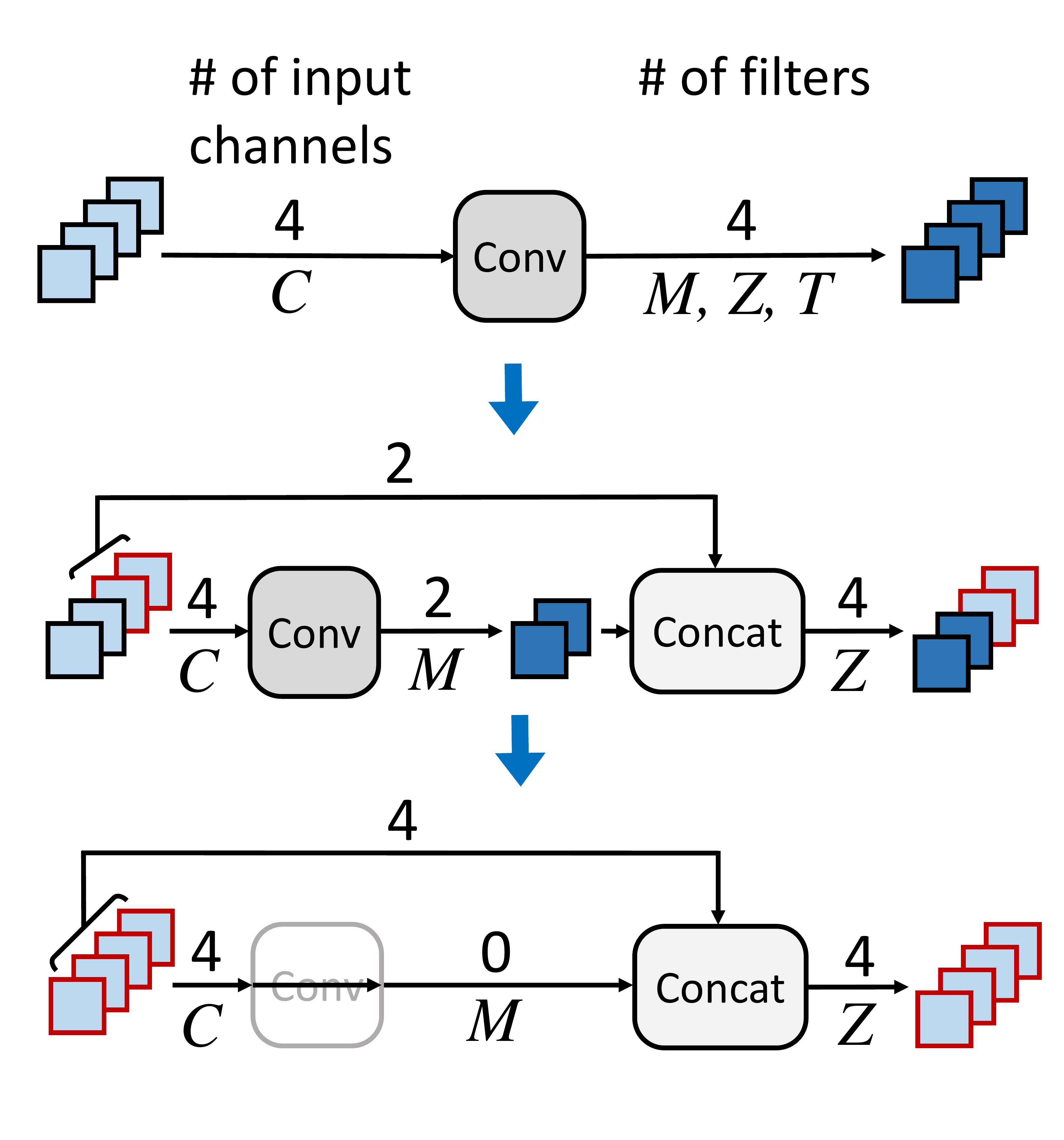}
        \label{fig:cbc_case1}
        
    }
    \hfill
    \centering
    \subfloat[Case 2: Fewer input channels than initial filters. \\ ($C = 4 < T = 6$)]
    {
        \includegraphics[width=0.225\linewidth]{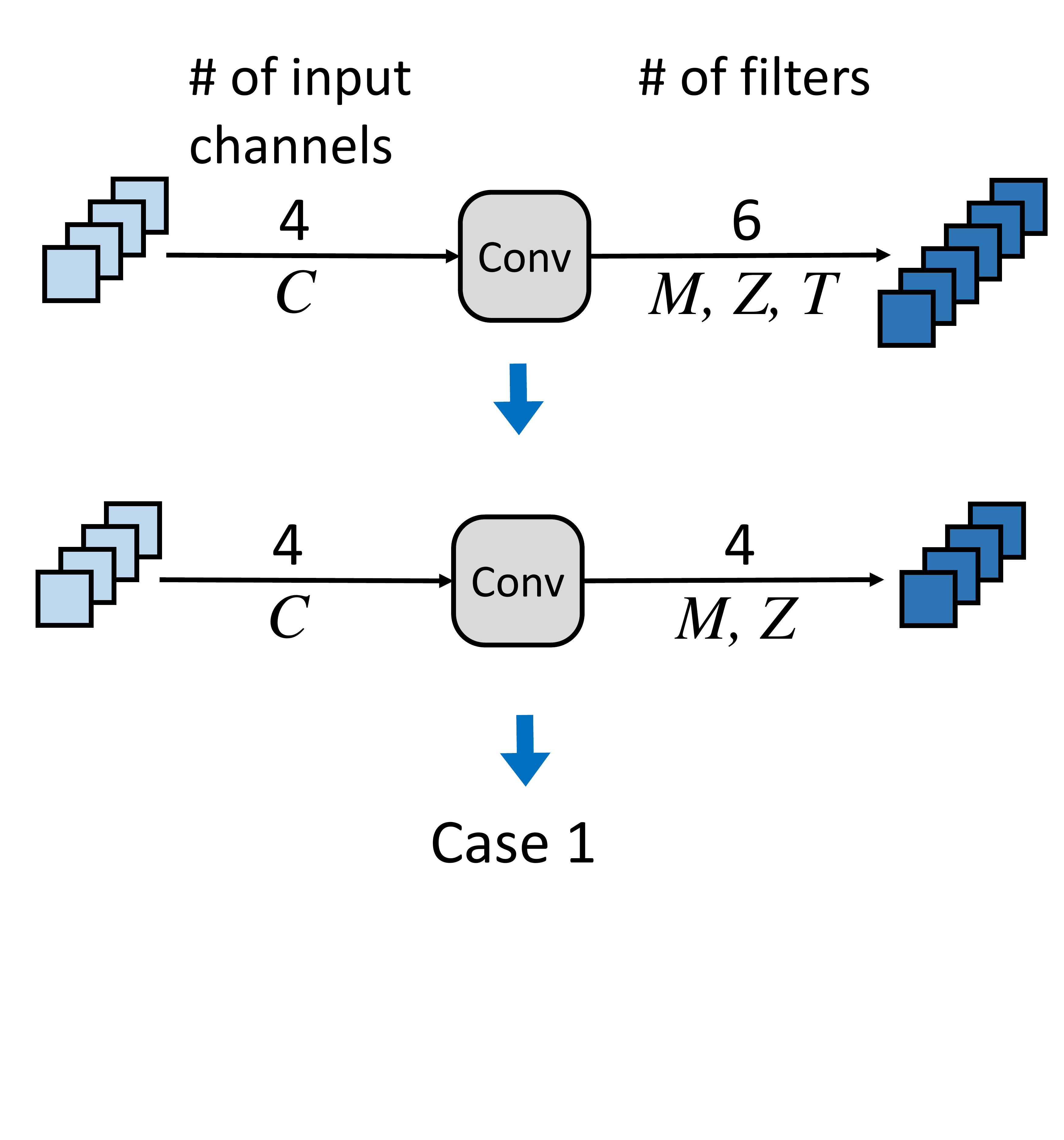}
        \label{fig:cbc_case2}
    }
    \hfill
    \centering
    \subfloat[Case 3: More input channels than initial filters. \\ ($C = 4 > T = 2$)]
    {
        \includegraphics[width=0.225\linewidth]{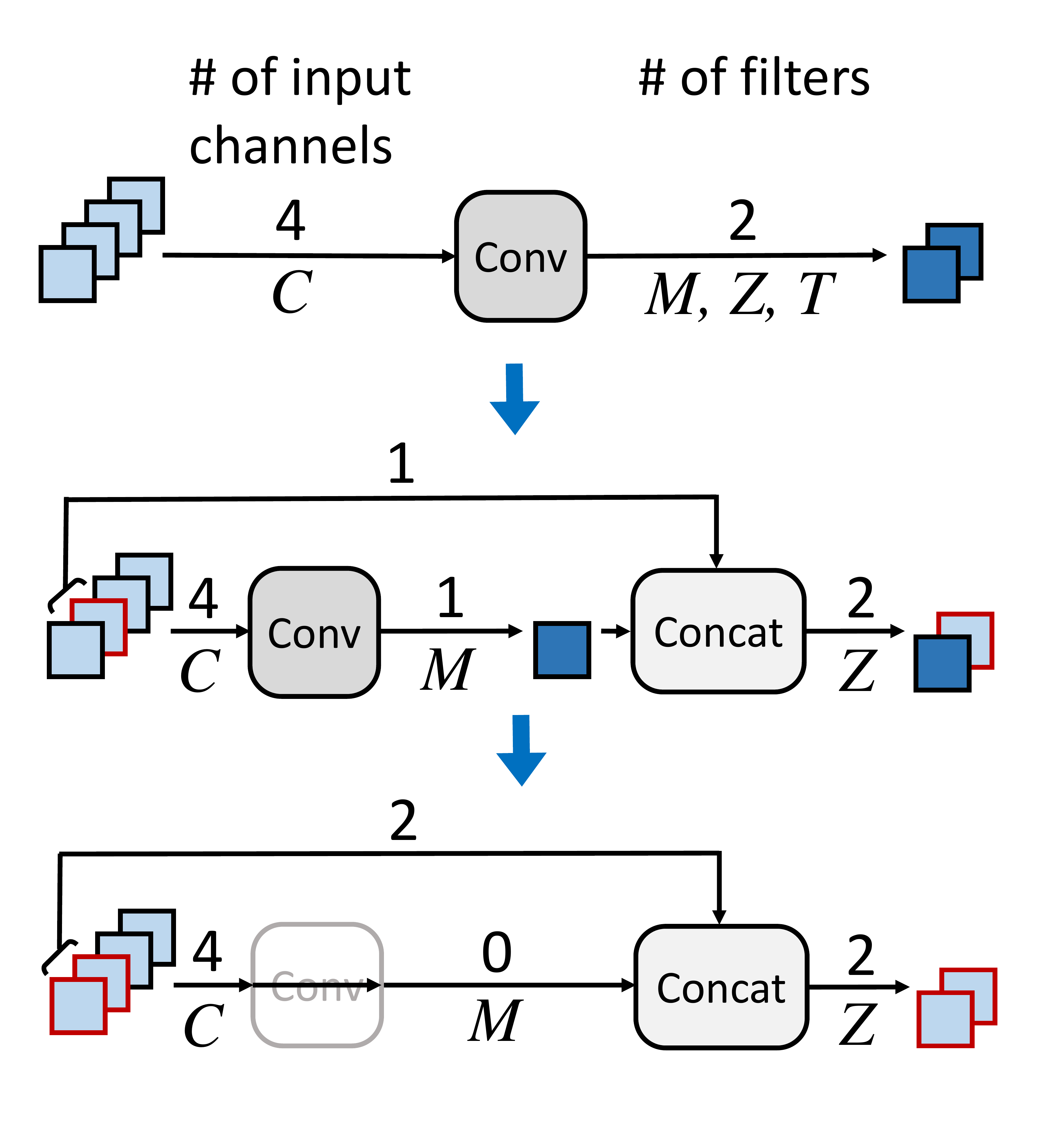}
        \label{fig:cbc_case3}
    }
    \hfill
    \centering
    \subfloat[Case 4: Same number of input channels and initial filters but with a given T. ($C = 4, T = 2$)]
    {
        \includegraphics[width=0.225\linewidth]{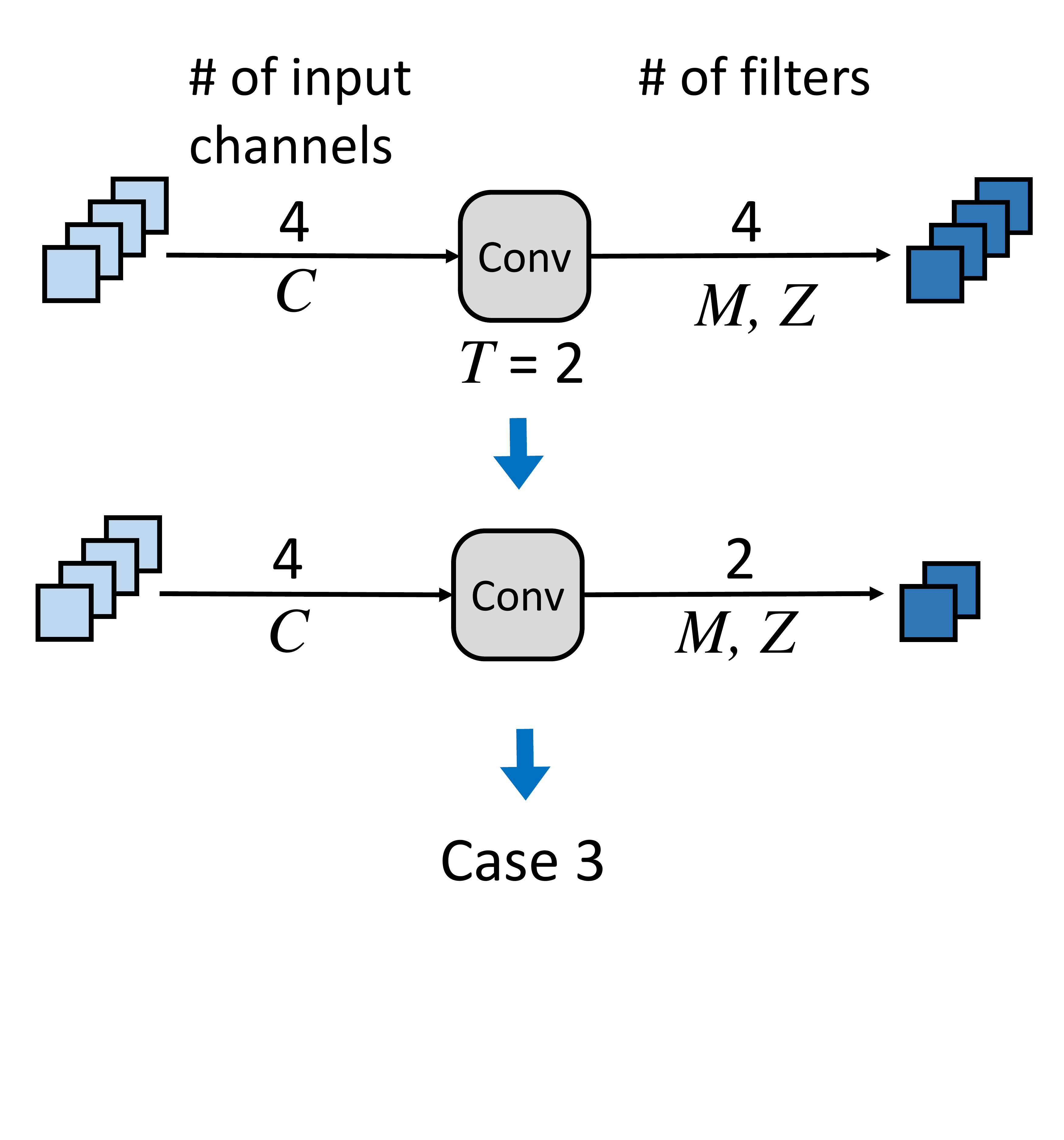}
        \label{fig:cbc_case4}
    }
    \caption{An illustration of how CBCs handle different cases based on the relationship between the number of input channels ($C$) and the initial number of filters ($T$) (only the number of filters remaining ($M$) changes, and $C$ and $T$ are fixed). For each case, it shows how the architecture changes with more filters removed from top to bottom. The numbers above lines correspond to the letters below lines. Please note that the number of output channels ($Z$) will never become zero.}
    \label{fig:cbc}
\end{figure*}

The high-level concept of CBCs is ``when a filter is removed, an input channel is bypassed to maintain the same number of output channels''. In this way, we can preserve the network connectivity when all filters are removed from a layer. Assuming the target layer in the initial network has $C$ input channels, $T$ filters, and $Z$ output channels\footnote{If we do not use CBCs, $Z$ is equal to $T$.}, we gradually remove filters from the layer, where there are $M$ filters remaining. Fig.~\ref{fig:cbc} illustrates how CBCs handle three cases in this process based on the relationship between the number of input channels ($C$) and the initial number of filters ($T$) (only $M$ changes, and $C$ and $T$ are fixed):
\begin{itemize}
    \item Case 1, $C = T$ (Fig.~\ref{fig:cbc_case1}): When the $i$-th filter is removed, we bypass the $i$-th input channel, so the number of output channels ($Z$) can be kept the same. When all the filters are removed ($M = 0$), all the input channels are bypassed, which is the same as removing the layer.
    \item Case 2, $C < T$ (Fig.~\ref{fig:cbc_case2}): We do not bypass input channels at the beginning of filter removal because we have more filters than input channels (i.e., $M > C$) and there are no corresponding input channels to bypass. The bypass process starts when there are fewer filters than input channels ($M < C$), which becomes case 1.
    \item Case 3, $C > T$ (Fig.~\ref{fig:cbc_case3}): When the $i$-th filter is removed, we bypass the $i$-th input channel. The extra ($C-T$) input channels are not used for the bypass.
\end{itemize}

These three cases can be summarized in a rule: when the $i$-th filter is removed, the corresponding $i$-th input channel is bypassed if that input channel exists. Therefore, the number of output channels ($Z$) when using CBCs can be computed by $Z = max(min(C, T), M)$. The proposed CBCs can be efficiently trained when combined with the proposed ordered dropout, as discussed in Sec.~\ref{subsec:ordered_droput}.

As a more advanced usage of $T$, we can treat $T$ as a hyper-parameter. Please note that we only change $M$, and $C$ and $T$ are fixed. From the formulation $Z = max(min(C, T), M)$, we can observe that the function of $T$ is limiting the number of bypassed input channels and hence the minimum number of output channels ($Z$). If we set $T \geq C$ to allow all $C$ input channels to be bypassed, the formulation becomes $Z = max(C, M)$, and the minimum number of output channels is $C$. If we set $T < C$ to only allow $T$ input channels to be bypassed, the formulation becomes $Z = max(T, M)$, and the minimum number of output channels is $T$. 

Setting $T < C$ enables generating the bottleneck, where we have fewer output channels than input channels ($Z < C$). The bottleneck has been shown to be effective in improving the accuracy-efficiency (e.g., accuracy-latency) trade-offs in MobileNetV2~\cite{Sandler_2018_CVPR}/V3~\cite{Howard_2019_ICCV}. We take the case 1 as an example. In Fig.~\ref{fig:cbc_case1}, we can observe that the number of output channels is always $4$, which is the same as the number of input channels ($Z=C=4$) no matter how many filters are removed. Therefore, the bottleneck cannot be generated. In contrast, if we set $T$ to 2 as the case 4 in Fig.~\ref{fig:cbc_case4}, no input channels are bypassed until we remove the first two filters because $Z = max(min(4, 2), 2) = 2$. After that, it becomes the case 3 in Fig.~\ref{fig:cbc_case3}, which forms a bottleneck.

\subsection{Ordered Dropout}
\label{subsec:ordered_droput}

Training the super-network involves joint training of multiple sub-networks with shared weights. After the super-network is trained, comparing sub-networks of the super-network (i.e., samples) only requires their \emph{relative} accuracy (e.g., sub-network A has higher accuracy than sub-network B). Generally speaking, the more sub-networks are trained, the better the relative accuracy of sub-networks will be. However, previous works usually require one forward-backward pass for training one sub-network. As a result, training more sub-networks requires more forward-backward passes and hence increases the training time.

\begin{figure}[t]
\begin{center}
   \includegraphics[width=0.9\linewidth]{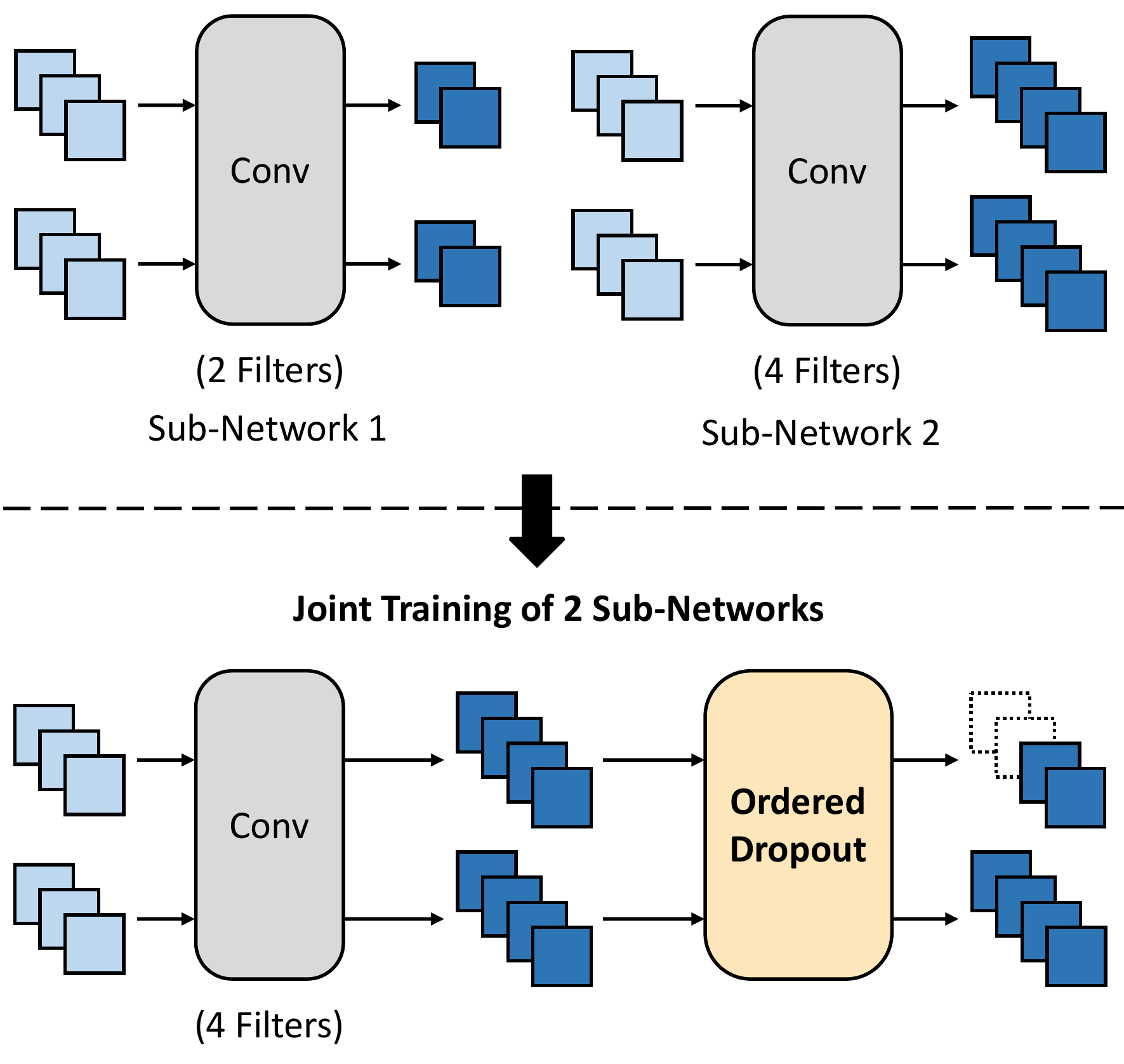}
\end{center}
   \caption{An illustration of how NetAdaptV2 uses the proposed ordered dropout to train two different sub-networks in a single forward-backward pass. The ordered dropout is inserted after each convolutional layer to simulate different layer widths by zeroing out some channels of activations. Note that all the sub-networks share the same set of weights.}
\label{fig:ordered_dropout}
\end{figure}

To address this problem, we propose ordered dropout (OD) to enable training $N$ sub-networks in a single forward-backward pass with a batch of $N$ images. OD is inserted after each convolutional layer in the super-network and zeros out \emph{different output channels for different images in a batch}. As shown in Fig.~\ref{fig:ordered_dropout}, OD simulates different layer widths with a constant number of output channels. Unlike the standard dropout~\cite{dropout} that zeros out a random subset of channels regardless of their positions, OD always zeros out the last channels to simulate removing the last filters. As a result, while sampling the search space, we can simply drop the last filters from the super-network to evaluate samples without other operations like sorting and avoid a mismatch between training and evaluation.

\begin{figure}[t]
\begin{center}
   \includegraphics[width=1.0\linewidth]{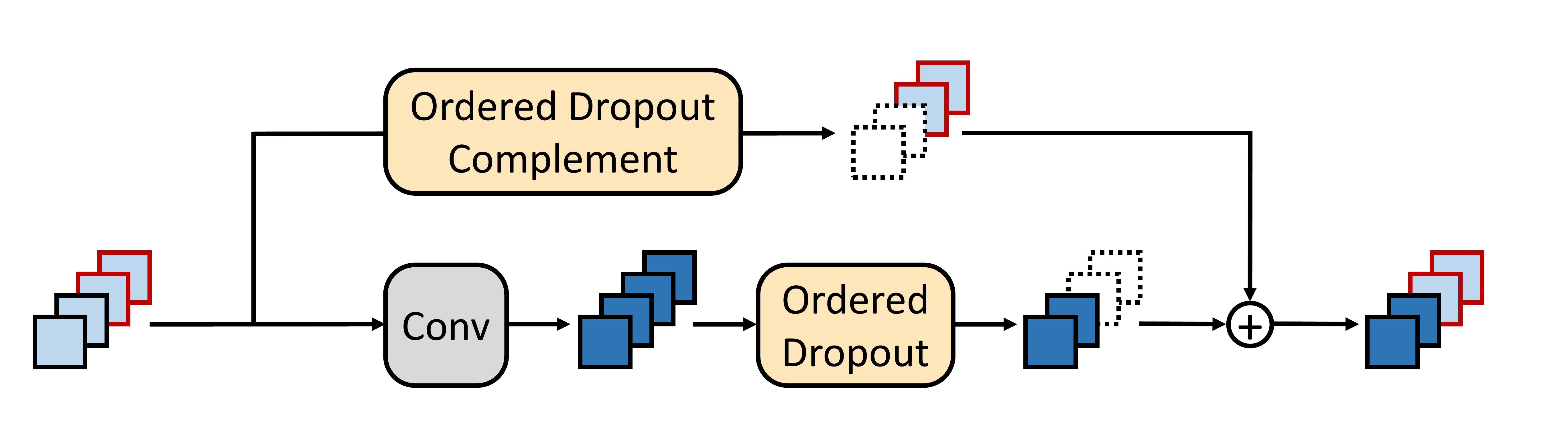}
\end{center}
   \vspace{-3mm}
   \caption{An illustration of how NetAdaptV2 uses the proposed channel-level bypass connections and ordered dropout to train a super-network that supports searching different layer widths and network depths.}
\label{fig:bypass_connections_train}
\end{figure}

When combined with the proposed CBCs, OD can train sub-networks with different network depths by zeroing out \emph{all} output channels of some layers to simulate layer removal. As shown in Fig.~\ref{fig:bypass_connections_train}, to simulate CBCs, there is another OD in the bypass path (upper) during training, which zeros out the \textit{complement} set of the channels zeroed by the OD in the corresponding convolution path (lower).

Because NAS only requires the relative accuracy of samples, we can decrease the number of training iterations to further reduce the super-network training time. Moreover, for each layer, we sample each layer width almost the same number of times in a forward-backward pass to avoid biasing towards any specific layer widths.

\subsection{Multi-Layer Coordinate Descent Optimizer}
\label{subsec:optimizer}

The single-layer coordinate descent (SCD) optimizer~\cite{book2020sze}, used in NetAdaptV1~\cite{eccv2018-netadapt}, is a simple-yet-effective optimizer with the advantages such as supporting both differentiable and non-differentiable search metrics and having only a few interpretable hyper-parameters that need to be tuned. The SCD optimizer runs an iterative optimization. It starts from the super-network and gradually reduces its latency (or other search metrics such as multiply-accumulate operations and energy). In each iteration, the SCD optimizer generates $K$ samples if the super-network has $K$ layers. The $k$-th sample is generated by shrinking (e.g., removing filters) the $k$-th layer in the best sample from the previous iteration to reduce its latency by a given amount. This amount is referred to as \emph{per-iteration resource reduction} and may change from one iteration to another. Then, the sample with the best performance (e.g., accuracy-latency trade-off) will be chosen and used for the next iteration. The optimization terminates when the target latency is met, and the sample with the best performance in the last iteration is the discovered DNN.

The shortcoming of the SCD optimizer is that it generates samples by shrinking only one layer per iteration. This property causes two problems. First, it does not consider the interplay of multiple layers when generating samples in an iteration, which may lead to sub-optimal performance of discovered DNNs. Second, it may take many iterations to search for very deep networks because the layer with the lowest latency limits the maximum value of the per-iteration resource reduction; the lowest latency of a layer becomes small when the super-network is deep. To address these problems, we propose the \emph{multi-layer coordinate descent (MCD) optimizer}. It generates $J$ samples per iteration, where each sample is obtained by randomly shrinking $L$ layers from the previous best sample. In NetAdaptV2, shrinking a layer involves removing filters, reducing the kernel size, or both. Compared with the SCD optimizer, the MCD optimizer considers the interplay of $L$ layers in each iteration so that the performance of the discovered DNN can be improved. Moreover, it enables using a larger per-iteration resource reduction (i.e., up to the total latency of $L$ layers) to reduce the number of iterations and hence the search time.

\section{Related Works}
\label{sec:related_works}

Reinforcement-learning-based methods~\cite{zoph2017nasreinforcement, zoph2018nasnet, Tan2018MnasNetPN, Chen2020MnasFPNLL, Tan2019EfficientNet} demonstrate the ability of neural architecture search for designing high-performance DNNs. However, its search time is longer than the following works due to the long time for training samples individually. Gradient-based methods~\cite{gordon2018morphnet, liu2018darts, wu2018fbnet, cai2018proxylessnas, stamoulis2019singlepath, stamoulis2019singlepathautoml, Mei2020AtomNAS, Xu2020PC-DARTS} successfully discover high-performance DNNs with a much shorter search time, but they can only support differentiable search metrics. NetAdaptV1~\cite{eccv2018-netadapt} proposes a single-layer coordinate descent optimizer that can support both differentiable and non-differentiable search metrics and was used to design state-of-the-art MobileNetV3~\cite{Howard_2019_ICCV}. However, shrinking only one layer for generating each sample and the long time for training samples individually become its bottleneck of search time. The idea of super-network training~\cite{autoslim_arxiv, cai2020once, yu2020bignas}, which jointly trains all the sub-networks in the search space, is proposed to reduce the time for training and evaluating samples and training the discovered DNN at the cost of a significant increase in the time for training a super-network. Moreover, network depth and layer width are usually considered separately in related works. The proposed NetAdaptV2 addresses all these problems at the same time by reducing the time for training a super-network, training and evaluating samples, and training the discovered DNN in balance while supporting non-differentiable search metrics.

The algorithm flow of NetAdaptV2 is most similar to NetAdaptV1~\cite{eccv2018-netadapt}, as shown in Fig.~\ref{fig:overview}. Compared with NetAdaptV2, NetAdaptV1 does not train a super-network but train each sample individually. Moreover, NetAdaptV1 considers only one layer per optimization iteration and different layer widths, but NetAdaptV2 considers multiple layers per optimization iteration and different layer widths, network depths, and kernel sizes. Therefore, NetAdaptV2 is both faster and more effective than NetAdaptV1, as shown in Sec.~\ref{subsubsec:ablation_study} and~\ref{subsec:depth_estimation}.

For the methodology, the proposed ordered dropout is most similar to the partial channel connections~\cite{Xu2020PC-DARTS}. However, they are different in the purpose and the ability to expand the search space. Partial channel connections aim to reduce memory consumption while training a DNN with multiple parallel paths by removing some channels. The number of channels removed is constant during training. Moreover, this number is manually chosen. As a result, partial channel connections do not expand the search space. In contrast, the proposed ordered dropout is designed for jointly training multiple sub-networks and expanding the search space. The number of channels removed (i.e., zeroed out) varies from image to image and from one training iteration to another during training to simulate different sub-networks. Moreover, the final number of channels removed (i.e., the discovered architecture) is searched. Therefore, the proposed ordered dropout expands the search space in terms of layer width as well as network depth when the proposed channel-level bypass connections are used.

\section{Experiment Results}

We apply NetAdaptV2 on two applications (image classification and depth estimation) and two search metrics (latency and multiply-accumulate operations (MACs)) to demonstrate the effectiveness and versatility of NetAdaptV2 across different operating conditions. We also perform an ablation study to show the impact of each of the proposed techniques and the associated hyper-parameters.

\subsection{Image Classification}
\label{sec:image_classification}

\subsubsection{Experiment Setup}
\label{subsec:experiment_setup}

We use latency or MACs to guide NetAdaptV2. The latency is measured on a Google Pixel 1 CPU. The search time is reported in GPU-hours and measured on V100 GPUs.

The dataset is ImageNet~\cite{ijcv2015-russakovsky-ilsvrc}. We reserve 10K images in the training set for comparing the accuracy of samples and train the super-network with the rest of the training images. The accuracy of the discovered DNN is reported on the validation set, which was not seen during the search. The initial network is based on MobileNetV3~\cite{Howard_2019_ICCV}. The maximum learning rate is 0.064 decayed by 0.963 every 3 epochs when the batch size is 1024. The learning rate scales linearly with the batch size~\cite{Goyal2017AccurateLM}. The optimizer is RMSProp with an $\ell$2 weight decay of $10^{-5}$. The dropout rate is 0.2. The decay rate of the exponential moving average is 0.9998. The batch size is 1024 for training the super-network, 2048 for training the latency-guided discovered DNN, and 1536 for training the MAC-guided discovered DNN.

The multi-layer coordinate descent (MCD) optimizer generates 200 samples per iteration ($J=200$). For the latency-guided experiment, each sample is obtained by randomly shrinking 10 layers ($L=10$) from the best sample in the previous iteration. We reduce the latency by 3\% in the first iteration (i.e., initial resource reduction) and decay the resource reduction by 0.98 every iteration. For the MAC-guided experiment, each sample is obtained by randomly shrinking 15 layers ($L=15$) from the best sample in the previous iteration. We reduce the MACs by 2.5\% in the first iteration and decay the resource reduction by 0.98 every iteration. More details are included in the appendix.

\subsubsection{Latency-Guided Search Result}
\label{subsec:latency_guided_search_result}

The results of NetAdaptV2 guided by latency and related works are summarized in Table~\ref{tab:nas_result}. Compared with the state-of-the-art (SOTA) NAS algorithms~\cite{cai2020once, yu2020bignas}, NetAdaptV2 reduces the search time by up to 5.8$\times$ and discovers DNNs with better accuracy-latency/accuracy-MAC trade-offs. The reduced search time is the result of the much more balanced time spent per step. Compared with the NAS algorithms in the class of hundreds of GPU-hours, ProxylessNAS~\cite{cai2018proxylessnas} and Single-Path NAS~\cite{stamoulis2019singlepathautoml}, NetAdaptV2 outperforms them without sacrificing the support of non-differentiable search metrics. NetAdaptV2 achieves either 2.4\% higher accuracy with $1.5\times$ lower latency or 1.4\% higher accuracy with $1.6\times$ lower latency. Compared with SOTA NAS-discovered MobileNetV3~\cite{Howard_2019_ICCV}, NetAdaptV2 achieves 1.8\% higher accuracy with the same latency in around 50 hours on eight GPUs. We estimate the $CO_2$ emission of NetAdaptV2 based on~\cite{strubell_2019_energy}. NetAdaptV2 discovers DNNs with better accuracy-latency/accuracy-MAC trade-offs with low $CO_2$ emission.

\begin{table*}[ht]
\centering
\scalebox{0.91}{
\begin{threeparttable}
\begin{tabular}{l|c|c|c|c|c|c}
\toprule
\multicolumn{1}{c|}{\multirow{2}{*}{Method}} & \multirow{2}{*}{\begin{tabular}[c]{@{}c@{}}Top-1 \\ Accuracy (\%)\end{tabular}} & \multirow{2}{*}{\begin{tabular}[c]{@{}c@{}}Latency \\ (ms)\end{tabular}} & \multirow{2}{*}{\begin{tabular}[c]{@{}c@{}}MAC\\ (M)\end{tabular}} & \multirow{2}{*}{\begin{tabular}[c]{@{}c@{}}Search Time\\ (GPU-Hours)\end{tabular}} & \multirow{2}{*}{\begin{tabular}[c]{@{}c@{}}Non-Diff. \\ Metrics \end{tabular}} & \multirow{2}{*}{\begin{tabular}[c]{@{}c@{}}$CO_2$ \\ Emission (lbs) \end{tabular}}
\\
\multicolumn{1}{c|}{}                        &                                                                                 &                               &                                                                    &      &                                                                              \\ \toprule
MnasNet~\cite{Tan2018MnasNetPN}                                       & 75.2                                                                        & 78            & 312        &                           -                              & \checkmark & -                                                                                \\ \hline
ProxylessNAS~\cite{cai2018proxylessnas}                                  & 74.6                                                                            & 78                             & 320                                                                & 500& & 142                                                                                \\ \hline
Single-Path NAS~\cite{stamoulis2019singlepathautoml}                                   & 75.6                                                                           & 82                            & -                                                                & \begin{tabular}[c]{@{}c@{}}243\tnote{*} \\ (24 (TPU V3), 0, 219)\end{tabular} &                 & 69                                                             \\ 
\hline
AutoSlim~\cite{autoslim_arxiv}                                      & 74.6                                                                            & 71                            & 315                                                                & -  & \checkmark & -                                                                               \\ \hline
FBNet~\cite{wu2018fbnet}                                         & 74.9                                                                            & -                             & 375                                                                & -     &        & -                                                                     \\ \hline
MobileNetV3~\cite{Howard_2019_ICCV}                                   & 75.2                                                                            & 51                            & 219                                                                & - & \checkmark       & -                                                                           \\ \hline

FairNAS~\cite{chu2019fairnas} & 76.7 & 77 & 325 & - & \checkmark & - \\ \hline

Once-for-All~\cite{cai2020once}                                   & 76.9                                                                           & 58                            & 230                                                                & \begin{tabular}[c]{@{}c@{}}1315 \\ (1200, 40, 75)\end{tabular} & \checkmark                & 374                                                             \\ \hline

BigNAS~\cite{yu2020bignas}                                   & 76.5                                                                           & -                            & 242                                                                & \begin{tabular}[c]{@{}c@{}}2304 (TPU V3) \\ (2304, -, 0)\end{tabular} & \checkmark                & 655                                                             \\ 
\hline                                    
\hline

\textbf{\begin{tabular}[l]{@{}l@{}}NetAdaptV2 \\ (Guided by Latency)\end{tabular}}                                   & \textbf{77.0}                                                                           & \textbf{51}                            & \textbf{225}                                                                & \textbf{\begin{tabular}[c]{@{}c@{}} 397 \\ (167, 24, 206)\end{tabular}} & \checkmark                & \textbf{113}                                                             \\ \toprule
\end{tabular}
\begin{tablenotes}\small
\item[*] 1) We merge the time for training the super-network and that for training and evaluating samples into one. 2) We train the discovered network for 350 epochs as mentioned in~\cite{stamoulis2019singlepathautoml}.
\end{tablenotes}
\end{threeparttable}
}
\caption{The comparison between NetAdaptV2 \emph{guided by latency} and related works on ImageNet. The number of MACs is reported for completeness although NetAdaptV2 is not guided by MACs and achieves sub-optimal accuracy-MAC trade-offs. The numbers between parentheses show the breakdown of the search time in terms of training a super-network, training and evaluating samples, and training the discovered DNN from left to right. \textit{Non-Diff. Metrics} denotes whether the method supports non-differentiable metrics. The last column \textit{$CO_2$ Emission} shows the estimated $CO_2$ emission based on~\cite{strubell_2019_energy}.}
\label{tab:nas_result}
\end{table*}

\subsubsection{MAC-Guided Search Result}
\label{subsec:mac_guided_search_result}

We present the result of NetAdaptV2 guided by MACs and compare it with related works in Table~\ref{tab:nas_result_large}. For a fair comparison, AutoAugment~\cite{autoaugment} and stochastic depth~\cite{stochastic_depth} with a survival probability of 0.8 are used for training the discovered network, which results in a longer time for training the discovered DNN. NetAdaptV2 achieves comparable accuracy-MAC trade-offs to NSGANetV2-m~\cite{lu2020nsganetv2} while the search time is $2.6 \times$ lower. Moreover, the discovered DNN also outperforms EfficientNet-B0~\cite{Tan2019EfficientNet} and MixNet-M~\cite{Tan2019MixConvMD} by up to $1.5\%$ higher top-1 accuracy with fewer MACs.

\begin{table}[t]
\scalebox{0.85}{
\begin{tabular}{l|c|c|c}
\toprule
\multicolumn{1}{c|}{Method} & \begin{tabular}[c]{@{}c@{}}Top-1\\ Accuracy (\%)\end{tabular} & \begin{tabular}[c]{@{}c@{}}MAC\\ (M)\end{tabular} & \begin{tabular}[c]{@{}c@{}}Search Time\\ (GPU-Hours)\end{tabular} \\ 
\toprule
NSGANetV2-m~\cite{lu2020nsganetv2}                  & 78.3                                                          & 312                                               & \begin{tabular}[c]{@{}c@{}}1674\\ (1200, 24, 450)\end{tabular}    \\ \hline
EfficientNet-B0~\cite{Tan2019EfficientNet}              & 77.3                                                          & 390                                               & -                                                                 \\ \hline
MixNet-M~\cite{Tan2019MixConvMD}                     & 77.0                                                          & 360                                               & -                                                                 \\ \hline \hline
\textbf{\begin{tabular}[c]{@{}l@{}}NetAdaptV2\\ (Guided by MAC)\end{tabular}} & \textbf{78.5}                                                 & \textbf{314}                                      & \textbf{\begin{tabular}[c]{@{}c@{}} 656 \\ (204, 35, 417)\end{tabular}}  \\ 
\toprule
\end{tabular}
}
\caption{The comparison between NetAdaptV2 \emph{guided by MACs} and related works. The numbers between parentheses show the breakdown of the search time in terms of training a super-network, training and evaluating samples, and training the discovered DNN from left to right.}
\label{tab:nas_result_large}
\end{table}

\subsubsection{Ablation Study}
\label{subsubsec:ablation_study}

The ablation study employs the experiment setup outlined in Sec.~\ref{subsec:experiment_setup} unless otherwise stated. To speed up training the discovered networks, the distillation model is smaller.

\noindent \textbullet\ \textbf{Impact of Ordered Dropout}

To study the impact of the proposed ordered dropout (OD), we do not use channel-level bypass connections (CBCs) and multi-layer coordinate descent (MCD) optimizer in this experiment. When we further remove the usage of OD, NetAdaptV2 becomes the same as NetAdaptV1~\cite{eccv2018-netadapt}, where each sample needs to be trained for four epochs by following the setting of NetAdaptV1. To speed up the execution of NetAdaptV1, we use a shallower network, MobileNetV1~\cite{Howard2017MobileNetV1}, in this experiment instead. Table~\ref{tab:ablation_ordered_dropout} shows that using OD reduces the search time by $3.3\times$ while achieving the same accuracy-latency trade-off. If we only consider the time for training a super-network and training and evaluating samples, which are affected by OD, the time reduction is $10.4\times$.

\begin{table}[t]
\centering
\scalebox{0.9}{
\begin{tabular}{c||c|c|c}
\toprule
OD & \begin{tabular}[c]{@{}c@{}}Top-1\\ Accuracy (\%)\end{tabular} & \begin{tabular}[c]{@{}c@{}}Latency\\ (ms)\end{tabular} & \begin{tabular}[c]{@{}c@{}}Search Time\\ (GPU-Hours)\end{tabular} \\ \toprule
& 71.0 (+0)                                                     & 43.9 (100\%)                                           & \begin{tabular}[c]{@{}c@{}}721 (100\%) \\ (0, 543, 178)\end{tabular}                                                      \\ \hline
\checkmark  & 71.1 (+0.1)                                                   & 44.4 (101\%)                                           & \begin{tabular}[c]{@{}c@{}}221 (31\%) \\ (50, 2, 169)\end{tabular}                                                         \\ \toprule
\end{tabular}
}
\caption{The ablation study of the proposed ordered dropout (OD) on MobileNetV1~\cite{Howard2017MobileNetV1} and ImageNet. The numbers between parentheses show the breakdown of the search time in terms of training a super-network, training and evaluating samples, and training the discovered DNN from left to right.}
\label{tab:ablation_ordered_dropout}
\end{table}

\noindent \textbullet\ \textbf{Impact of Channel-Level Bypass Connections}

The proposed channel-level bypass connections (CBCs) enable NetAdaptV2 to search for different network depths. Table~\ref{tab:ablation_cbc_optimizer} shows that CBCs can improve the accuracy by 0.3\%. The difference is more significant when we target at lower latency, as shown in the ablation study on MobileNetV1 in the appendix, because the ability to remove layers becomes more critical for maintaining accuracy.

\begin{table}[t]
\centering
\scalebox{0.9}{
\begin{tabular}{c|c||c}
\toprule
\multicolumn{2}{c||}{Methods} & \multirow{2}{*}{\begin{tabular}[c]{@{}c@{}}Top-1\\ Accuracy (\%)\end{tabular}} \\ \cline{1-2}
CBC           & MCD           &                                                                                \\ \hline
              &               & 75.9 (+0)                                                                           \\ \hline
\checkmark             &               & 76.2 (+0.3)                                                                           \\ \hline
\checkmark             & \checkmark             & 76.6 (+0.7)                                                                           \\ \toprule
\end{tabular}
}
\caption{The ablation study of the channel-level bypass connections (CBCs) and the multi-layer coordinate descent optimizer (MCD) on ImageNet. The latency of the discovered networks is around 51ms, and ordered dropout is used.}
\label{tab:ablation_cbc_optimizer}
\end{table}

\noindent \textbullet\ \textbf{Impact of Multi-Layer Coordinate Descent Optimizer}

The proposed multi-layer coordinate descent (MCD) optimizer improves the performance of the discovered DNN by considering the joint effect of multiple layers per optimization iteration. Table~\ref{tab:ablation_cbc_optimizer} shows that using the MCD optimizer further improves the accuracy by 0.4\%.

\noindent \textbullet\ \textbf{Impact of Resource Reduction and Number of Samples}

The two main hyper-parameters of the MCD optimizer are the per-iteration resource reduction, which is defined by an initial resource reduction and a decay rate, and the number of samples per iteration ($J$). They influence the accuracy of the discovered networks and the search time. Table~\ref{tab:influence_latency_reduction} summarizes the accuracy of the 51ms discovered networks when using different initial latency reductions (with a fixed decay of 0.98 per iteration) and different numbers of samples.

The first experiment is fixing the number of samples per iteration and increasing the initial latency reduction from 1.5\% to 6.0\%, which gradually reduces the time for evaluating samples. The result shows that as long as the latency reduction is small enough, specifically below 3\% in this experiment, the accuracy of the discovered networks does not change with the latency reduction.

The second experiment is fixing the time for evaluating samples by scaling both the initial latency reduction and the number of samples per iteration at the same rate. As shown in Table~\ref{tab:influence_latency_reduction}, as long as the latency reduction is small enough, more samples will result in better discovered networks. However, if the initial latency reduction is too large, increasing the number of samples per iteration cannot prevent the accuracy from degrading.

\begin{table}[t]
\centering
\scalebox{0.8}{
\begin{tabular}{c||c|c|c}
\toprule
                                                                                             \multicolumn{1}{c||}{} & \multicolumn{1}{c|}{\begin{tabular}[c]{@{}c@{}}Initial Latency\\ Reduction\end{tabular}} & \multicolumn{1}{c|}{\begin{tabular}[c]{@{}c@{}}Number of \\ Samples ($J$)\end{tabular}} & \multicolumn{1}{c}{\begin{tabular}[c]{@{}c@{}}Top-1\\ Accuracy (\%)\end{tabular}} \\ \toprule
\multirow{3}{*}{\begin{tabular}[c]{@{}c@{}}Fixed Number of\\ Samples per Iteration\end{tabular}}  & 1.5\%                                                               & 100                                                                      & 76.4   \\ \cline{2-4} 
                                                                                             & 3.0\%                                                               & 100                                                                      & 76.4   \\ \cline{2-4} 
                                                                                             & 6.0\%                                                               & 100                                                                      & 75.9   \\ \toprule
\multirow{3}{*}{\begin{tabular}[c]{@{}c@{}}Fixed Time for\\ Evaluating Samples\end{tabular}} & 1.5\%                                                               & 100                                                                      & 76.4   \\ \cline{2-4} 
                                                                                             & 3.0\%                                                               & 200                                                                      & 76.6   \\ \cline{2-4} 
                                                                                             & 6.0\%                                                               & 400                                                                      & 75.7   \\ \toprule
\end{tabular}
}
\caption{The experiments for evaluating the influence of the two main hyper-parameters of the MCD optimizer, which are the initial latency reduction (with a fixed decay of 0.98 per iteration) and the number of samples ($J$). All discovered networks have almost the same latency (51ms).}
\label{tab:influence_latency_reduction}
\end{table}

\noindent \textbullet\ \textbf{Accuracy Variation across Multiple Executions}

To know the accuracy variation of each step in NetAdaptV2~\cite{li2020reproducibility}, we execute different steps three times and summarize the resultant accuracy of the discovered networks in Table~\ref{tab:reproducibility}. The initial latency reduction is 1.5\%, and the number of samples per iteration is 100 ($J=100$). The latency of discovered networks is around 51ms. According to the last row of Table~\ref{tab:reproducibility}, which corresponds to executing the entire algorithm flow of NetAdaptV2 three times, the accuracy variation is 0.3\%. The variation is fairly small because simply training the same discovered network three times results in an accuracy variation of 0.1\% as shown in the first row. Moreover, when we fix the super-network and execute the MCD optimizer three times as shown in the second row, the accuracy variation is the same as that of executing the entire NetAdaptV2 three times. The result suggests that the randomness in training a super-network does not increase the overall accuracy variation, which is preferable since we only need to perform this relatively costly step one time.

\begin{table}[t]
\centering
\scalebox{0.8}{
\begin{tabular}{c|c|c||c|c|c}
\toprule
\multirow{2}{*}{\begin{tabular}[c]{@{}c@{}}Training\\ Super-Network\end{tabular}} & \multirow{2}{*}{\begin{tabular}[c]{@{}c@{}}Evaluating\\ Samples\end{tabular}} & \multirow{2}{*}{\begin{tabular}[c]{@{}c@{}}Training\\ Discovered DNN\end{tabular}} & \multicolumn{3}{c}{\begin{tabular}[c]{@{}c@{}}Top-1 Accuracy of\\ Executions (\%)\end{tabular}} \\ \cline{4-6}  
                                                                                     &                                                                               &                                                                                    & 1       & 2       & 3      \\ \toprule
                                                                                     &                                                                              & \checkmark                                                                                  & 76.1    & 76.1    & 76.2   \\  \hline
                                                                                     & \checkmark                                                                             & \checkmark                                                                                  & 76.1    & 76.2    & 76.4   \\  \hline
\checkmark                                                                                    & \checkmark                                                                             & \checkmark                                                                                  & 76.1    & 76.2    & 76.4   \\ \toprule
\end{tabular}
}
\caption{The accuracy variation of NetAdaptV2. The $\checkmark$ denotes the step is executed three times, and the others are executed once. For example, the last row corresponds to executing the entire algorithm flow of NetAdaptV2 three times. For the MCD optimizer, the initial latency reduction is 1.5\%, and the number of samples per iteration is 100 ($J=100$). The latency of all discovered networks is around 51ms, and the accuracy values are sorted in ascending order.}
\label{tab:reproducibility}
\end{table}

\subsection{Depth Estimation}
\label{subsec:depth_estimation}
\subsubsection{Experiment Setup}
\label{subsubsec:depth_estimation_experiment_setup}

NYU Depth V2~\cite{nyudepth} is used for depth estimation. We reserve 2K training images for evaluating the performance of samples and train the super-network with the rest of the training images. The initial network is FastDepth~\cite{icra_2019_fastdepth}.
Following FastDepth, we pre-train the encoder of the super-network on ImageNet.
The batch size is 256, and the learning rate is 0.9 decayed by 0.963 every epoch. After pre-training the encoder,  we train the super-network on NYU Depth V2 for 50 epochs with a batch size of 16 and an initial learning rate of 0.025 decayed by 0.9 every epoch. For the MCD optimizer, we generate 150 ($J=150$) samples per iteration. We search with latency measured on a Google Pixel 1 CPU. The latency reduction is 1.5\% in the first iteration and is decayed by 0.98 every iteration. For training the discovered network, we use the same setup as training the super-network, except that the initial learning rate is 0.05. 

\subsubsection{Search Result}

The comparison between the proposed NetAdaptV2 and NetAdaptV1~\cite{eccv2018-netadapt}, which is used in FastDepth~\cite{icra_2019_fastdepth}, is summarized in Table~\ref{tab:fastdepth}. NetAdaptV2 reduces the search time by $2.4\times$ on NYU Depth V2, and the discovered DNN outperforms that of NetAdaptV1 by 0.5\% in delta-1 accuracy with comparable latency. Because NYU Depth V2 is much smaller than ImageNet, the reduction in the total search time is less than that of applying NetAdaptV2 on ImageNet. The search time spent on ImageNet is for pre-training the encoder, which is a common practice and indispensable when training DNNs for depth estimation on NYU Depth V2.

\begin{table}[t]
\centering
\scalebox{0.8}{
\begin{tabular}{l|c|c|c|c|c}
\toprule
\multicolumn{1}{c|}{\multirow{2}{*}{Method}} & \multirow{2}{*}{\begin{tabular}[c]{@{}c@{}}RMSE \\ (m)\end{tabular}} & \multirow{2}{*}{\begin{tabular}[c]{@{}c@{}}Delta-1\\ Accuracy\\ (\%)\end{tabular}} & \multirow{2}{*}{\begin{tabular}[c]{@{}c@{}}Latency \\ (ms)\end{tabular}} & \multicolumn{2}{c}{\begin{tabular}[c]{@{}c@{}}Search Time\\ (GPU-Hours)\end{tabular}} \\ \cline{5-6} 
\multicolumn{1}{c|}{}                        &                           &                                                                                   &                               & ImageNet                                 & \begin{tabular}[c]{@{}c@{}}NYU \\ Depth\end{tabular}                                   \\ \toprule
NetAdaptV1~\cite{eccv2018-netadapt}                                      & 0.583                     & 77.4                                                                              & 87.6                          & 96                                       & 65                                          \\ \hline
\textbf{NetAdaptV2}                            & \textbf{0.576}            & \textbf{77.9}                                                                     & \textbf{86.7}                 & 96                                       & \textbf{27}                                 \\ \toprule
\end{tabular}
}
\caption{The comparison between NetAdaptV2 and NetAdaptV1 on depth estimation and NYU Depth V2~\cite{nyudepth}.}
\label{tab:fastdepth}
\end{table}

\section{Conclusion}

In this paper, we propose NetAdaptV2, an efficient neural architecture search algorithm, which significantly reduces the total search time and discovers DNNs with state-of-the-art accuracy-latency/accuracy-MAC trade-offs. 
NetAdaptV2 better balances the time spent per step and supports non-differentiable search metrics. This is realized by the proposed methods: channel-level bypass connections, ordered dropout, and multi-layer coordinate descent optimizer. The experiments demonstrate that NetAdaptV2 can reduce the total search time by up to $5.8\times$ on image classification and $2.4\times$ on depth estimation and discover DNNs with better performance than state-of-the-art works.

\section*{Acknowledgement}
This research was funded by the National Science Foundation, Real-Time Machine Learning (RTML) program, through grant No. 1937501, a Google Research Award, and gifts from Intel and Facebook.

\clearpage

\appendix
\section{Additional Information about Experiment Setup}
This section provides additional information about the experiment setup for image classification (Sec.~\ref{sec:image_classification}) and depth estimation (Sec.~\ref{subsec:depth_estimation}).

\subsection{Image Classification}

For the latency-guided experiment, we design the initial network based on MobileNetV3~\cite{Howard_2019_ICCV} with an input resolution of $224\times224$, which is widely used to construct the search space. Starting from MobileNetV3-Large, we round up all layer widths to the power of 2 and add two MobileNetV3 blocks, each with the input resolution of $28\times28$ and $14\times14$. The kernel sizes of depthwise layers are increased by 2. For sampling sub-networks, we allow 9 uniformly-spaced layer widths from 0 to the full layer width and different odd kernel sizes from 3 to the full kernel size for each layer. We use the initial network of Once-for-All~\cite{cai2020once} for knowledge distillation when training the discovered network for a fair comparison with Once-for-All~\cite{cai2020once} and BigNAS~\cite{yu2020bignas}.

For the MAC-guided experiment, following the practice of Once-for-All~\cite{cai2020once} and NSGANetV2~\cite{lu2020nsganetv2}, we increase layer widths of the initial network used in the latency-guided experiment by $1.25 \times$ and add one MobileNetV3 block with an input resolution of $7 \times 7$ to support a large-MAC operating condition. For sampling sub-networks, we allow 11 uniformly-spaced layer widths from 0 to the full layer width and different odd kernel sizes from 3 to the full kernel size for each layer. When training the discovered DNN, we use the initial network of Once-for-All~\cite{cai2020once} for knowledge distillation for a fair comparison with NSGANetV2~\cite{lu2020nsganetv2}.

\subsection{Depth Estimation}

The initial network is FastDepth~\cite{icra_2019_fastdepth}. FastDepth consists of an encoder and a decoder. The encoder uses MobileNetV1~\cite{Howard2017MobileNetV1} as a feature extractor, and the decoder uses depthwise separable convolution and nearest neighbor upsampling. Please refer to FastDepth~\cite{icra_2019_fastdepth} for more details. For sampling sub-networks, we allow 9 uniformly-spaced layer widths from 0 to the full layer width and different odd kernel sizes from 3 to the full kernel size for each layer.

Following a common practice of training DNNs on the NYU Depth V2 dataset, we pre-train the encoder of the initial network on ImageNet. However, as shown in Table~\ref{tab:fastdepth}, this step is relatively expensive and takes 96 GPU-hours. To avoid pre-training the encoder of the discovered DNN on ImageNet again, we transfer the knowledge learned by the encoder of the initial network to that of the discovered DNN, which is achieved by the following method. We log the architecture of the best sample in each iteration of the MCD optimizer, which forms an architecture trajectory. The starting point of this trajectory is the initial DNN architecture, and the end point is the discovered DNN architecture. For training the discovered network, we start from the starting point of the trajectory with the pre-trained weights of the initial DNN and follow this trajectory to gradually shrink the architecture. In each step of shrinking the architecture, we reuse the overlapped weights from the previous architecture and train the new architecture for two epochs. This process continues until we get to the end point of the trajectory, which is the discovered DNN architecture. Then, we train it until convergence. This knowledge transfer method enables high accuracy of the discovered DNN without pre-training its encoder on ImageNet.

\section{Discovered DNN Architectures}

For the latency-guided experiment on ImageNet (Table~\ref{tab:nas_result}), Table~\ref{tab:discovered_architecture} shows the discovered 51ms DNN architecture of NetAdaptV2. To make the numbers of MACs of all layers as similar as possible, modern DNN design usually doubles the number of filters and channels when the resolution of activations is reduced by 2$\times$. Similarly, to fix the ratio of $T$ (Sec.~\ref{subsec:channel_level_bypass_connections}) to the number of input channels, we use one value of $T$ for each resolution of input activations and set $T$ inversely proportional to the resolution. $T$s of all depthwise layers are set to infinity to allow bypassing all the input channels. We observe that channel-level bypass connections (CBCs) are widely used in the discovered DNN. Moreover, block 12 is removed, which demonstrates the ability of CBCs to remove a layer.

\begin{table*}[t]
\centering
\begin{tabular}{c|c|c|c|c|c|c|c|c|c|c}
\toprule
Index & Type            & $T$    & Kernel Size & Stride & BN      & Act & Exp  & DW   & PW   & SE  \\ \toprule
1     & conv            & 8    & 3                                                     & 2      & $\checkmark$ & HS  & 16   & -    & -    & -   \\ \hline
2     & mnv3 block      & 8    & 3                                                     & 1      & $\checkmark$ & RE  & 8    & 8    & 16   & -   \\ \hline
3     & mnv3 block      & 16   & 5                                                     & 2      & $\checkmark$ & RE  & 48   & 48   & 20   & 16  \\ \hline
4     & mnv3 block      & 16   & 3                                                     & 1      & $\checkmark$ & RE  & 48   & 48   & 32   & -   \\ \hline
5     & mnv3 block      & 32   & 7                                                     & 2      & $\checkmark$ & RE  & 80   & 80   & 32   & 32  \\ \hline
6     & mnv3 block      & 32   & 3                                                     & 1      & $\checkmark$ & RE  & 112  & 80   & 40   & 32  \\ \hline
7     & mnv3 block      & 32   & 3                                                     & 1      & $\checkmark$ & RE  & 64   & 32   & 16   & 32  \\ \hline
8     & mnv3 block      & 32   & 3                                                     & 1      & $\checkmark$ & RE  & 96   & 96   & 8    & 32  \\ \hline
9     & mnv3 block      & 64   & 5                                                     & 2      & $\checkmark$ & HS  & 192  & 192  & 128  & 64  \\ \hline
10    & mnv3 block      & 64   & 5                                                     & 1      & $\checkmark$ & HS  & 224  & 192  & 128  & -   \\ \hline
11    & mnv3 block      & 64   & 3                                                     & 1      & $\checkmark$ & HS  & 128  & 32   & 48   & 64  \\ \hline
12    & mnv3 block      & 64   & 0                                                     & 1      & $\checkmark$ & HS  & 0    & 0    & 0    & 0   \\ \hline
13    & mnv3 block      & 64   & 3                                                     & 1      & $\checkmark$ & HS  & 512  & 256  & 80   & 256 \\ \hline
14    & mnv3 block      & 64   & 3                                                     & 1      & $\checkmark$ & HS  & 256  & 256  & 112  & 256 \\ \hline
15    & mnv3 block      & 64   & 5                                                     & 1      & $\checkmark$ & HS  & 512  & 512  & 64   & 256 \\ \hline
16    & mnv3 block      & 128  & 7                                                     & 2      & $\checkmark$ & HS  & 640  & 640  & 224  & 256 \\ \hline
17    & mnv3 block      & 128  & 7                                                     & 1      & $\checkmark$ & HS  & 640  & 384  & 224  & 256 \\ \hline
18    & mnv3 block      & 128  & 5                                                     & 1      & $\checkmark$ & HS  & 896  & 512  & 224  & 256 \\ \hline
19    & conv            & 128  & 1                                                     & 1      & $\checkmark$ & HS  & 1024 & -    & -    & -   \\ \hline
20    & global avg pool & -    & -                                                     & -      & -            & -   & -    & -    & -    & -   \\ \hline
21    & conv            & 1024 & 1                                                     & 1      &              & HS  & 1792 & -    & -    & -   \\ \hline
22    & fc              & -    & 1                                                     & 1      & -            & -   & 1000 & -    & -    & -   \\ \toprule
\end{tabular}
\caption{The discovered 51ms DNN architecture of NetAdaptV2 on ImageNet presented in Table~\ref{tab:nas_result}. Type: type of the layer or block. BN: using batch normalization. Act: activation type (HS: Hard-Swish, RE: ReLU). Exp: number of filters in the expansion layer or number of filters in the conv layer. DW: number of filters in the depthwise layer. PW: number of filters in the pointwise layer. SE: number of filters in the squeeze-and-excitation operation. All MobileNetV3 blocks (mnv3 block) with a stride of 1 have residual connections.}
\label{tab:discovered_architecture}
\end{table*}

For the MAC-guided experiment on ImageNet (Table~\ref{tab:nas_result_large}), Table~\ref{tab:discovered_architecture_mac} shows the discovered 314M-MAC DNN architecture. We apply the same rule for setting $T$s as in the latency-guided experiment. We observe that CBCs are widely used in the discovered DNN. Moreover, MobileNetV3 block 7, 8, 12, 15 are removed.

\begin{table*}[t]
\centering
\begin{tabular}{c|c|c|c|c|c|c|c|c|c|c}
\toprule
Index & Type            & $T$    & Kernel Size & Stride & BN      & Act & Exp  & DW   & PW   & SE  \\ \toprule
1     & conv            & 8    & 3                                                     & 2      & $\checkmark$ & HS  & 24   & -    & -    & -   \\ \hline
2     & mnv3 block      & 8    & 3                                                     & 1      & $\checkmark$ & RE  & -    & 24    & 24   & -   \\ \hline

3     & mnv3 block      & 16   & 5                                                     & 2      & $\checkmark$ & RE  & 64   & 48   & 32   & 24  \\ \hline
4     & mnv3 block      & 16   & 3                                                     & 1      & $\checkmark$ & RE  & 128  & 64   & 32   & -   \\ \hline

5     & mnv3 block      & 32   & 5                                                     & 2      & $\checkmark$ & RE  & 96   & 96   & 48   & 40  \\ \hline
6     & mnv3 block      & 32   & 3                                                     & 1      & $\checkmark$ & RE  & 128 & 80   & 56   & 40  \\ \hline
7     & mnv3 block      & 32   & 0                                                     & 1      & $\checkmark$ & RE  & 0  & 0   & 0   & 0  \\ \hline
8     & mnv3 block      & 32   & 0                                                     & 1      & $\checkmark$ & RE  & 0   & 0   & 0   & 0  \\ \hline

9     & mnv3 block      & 64   & 5                                                     & 2      & $\checkmark$ & HS  & 224  & 224  & 96  & 80  \\ \hline
10    & mnv3 block      & 64   & 3                                                     & 1      & $\checkmark$ & HS  & 224  & 96  & 96  & -   \\ \hline
11    & mnv3 block      & 64   & 3                                                     & 1      & $\checkmark$ & HS  & 256  & 256   & 96   & 80  \\ \hline
12    & mnv3 block      & 64   & 0                                                     & 1      & $\checkmark$ & HS  & 0    & 0    & 0    & 0   \\ \hline

13    & mnv3 block      & 64   & 5                                                     & 1      & $\checkmark$ & HS  & 640  & 640  & 144   & 320 \\ \hline
14     & mnv3 block      & 64   & 3                                                     & 1      & $\checkmark$ & HS  & 640  & 512   & 144   & 320  \\ \hline
15    & mnv3 block      & 64   & 0                                                     & 1      & $\checkmark$ & HS  & 0  & 0  & 0  & 0 \\ \hline

16    & mnv3 block      & 128  & 5                                                     & 2      & $\checkmark$ & HS  & 768  & 640  & 192  & 320 \\ \hline
17    & mnv3 block      & 128  & 5                                                     & 1      & $\checkmark$ & HS  & 768  & 256  & 192  & 320 \\ \hline
18    & mnv3 block      & 128  & 7                                                     & 1      & $\checkmark$ & HS  & 896  & 768  & 192  & 320 \\ \hline
19    & mnv3 block      & 128  & 7                                                     & 1      & $\checkmark$ & HS  & 1152  & 1152  & 192  & 320 \\ \hline

20    & conv            & 128  & 1                                                     & 1      & $\checkmark$ & HS  & 1152 & -    & -    & -   \\ \hline

21    & global avg pool & -    & -                                                     & -      & -            & -   & -    & -    & -    & -   \\ \hline
22    & conv            & 1024 & 1                                                     & 1      &              & HS  & 2048 & -    & -    & -   \\ \hline
23    & fc              & -    & 1                                                     & 1      & -            & -   & 1000 & -    & -    & -   \\ \toprule
\end{tabular}
\caption{The discovered 314M-MAC DNN architecture of NetAdaptV2 on ImageNet presented in Table~\ref{tab:nas_result_large}. Type: type of the layer or block. BN: using batch normalization. Act: activation type (HS: Hard-Swish, RE: ReLU). Exp: number of filters in the expansion layer or number of filters in the conv layer. DW: number of filters in the depthwise layer. PW: number of filters in the pointwise layer. SE: number of filters in the squeeze-and-excitation operation. All MobileNetV3 blocks (mnv3 block) with a stride of 1 have residual connections.}
\label{tab:discovered_architecture_mac}
\end{table*}

For the depth estimation experiment on the NYU Depth V2 dataset (Table~\ref{tab:fastdepth}), Table~\ref{tab:discovered_architecture_fastdepth} shows the discovered 87ms DNN architecture of NetAdaptV2. We apply the same rule for setting $T$s as in the image classification experiments. We observe that NetAdaptV2 reduces the kernel sizes of the $37$-th and $40$-th depthwise convolutional layers from 5 to 3, which demonstrates that the ability to search kernel sizes may improve the performance of the discovered DNN.

\begin{table*}[t]
\centering
\begin{tabular}{c|c|c|c|c|c}
\toprule
Index & Type     & $T$   & Kernel Size & Stride & Filter  \\ \toprule
1     & conv     & 16  & 3           & 2      & 24  \\ \hline
2     & dw       & $\infty$ & 3           & 1      & 20  \\ \hline
3     & pw       & 16  & 1           & 1      & 48  \\ \hline
4     & dw       & $\infty$ & 3           & 2      & 48  \\ \hline
5     & pw       & 32  & 1           & 1      & 96  \\ \hline
6     & dw       & $\infty$ & 3           & 1      & 96  \\ \hline
7     & pw       & 32  & 1           & 1      & 112 \\ \hline
8     & dw       & $\infty$ & 3           & 2      & 112 \\ \hline
9     & pw       & 64  & 1           & 1      & 256 \\ \hline
10    & dw       & $\infty$ & 3           & 1      & 256 \\ \hline
11    & pw       & 64  & 1           & 1      & 192 \\ \hline
12    & dw       & $\infty$ & 3           & 2      & 192 \\ \hline
13    & pw       & 128 & 1           & 1      & 448 \\ \hline
14    & dw       & $\infty$ & 3           & 1      & 448 \\ \hline
15    & pw       & 128 & 1           & 1      & 448 \\ \hline
16    & dw       & $\infty$ & 3           & 1      & 448 \\ \hline
17    & pw       & 128 & 1           & 1      & 384 \\ \hline
18    & dw       & $\infty$ & 3           & 1      & 384 \\ \hline
19    & pw       & 128 & 1           & 1      & 512 \\ \hline
20    & dw       & $\infty$ & 3           & 1      & 512 \\ \hline
21    & pw       & 128 & 1           & 1      & 384 \\ \hline
22    & dw       & $\infty$ & 3           & 1      & 384 \\ \hline
23    & pw       & 128 & 1           & 1      & 448 \\ \hline
24    & dw       & $\infty$ & 3           & 2      & 448 \\ \hline
25    & pw       & 256 & 1           & 1      & 384 \\ \hline
26    & dw       & $\infty$ & 3           & 1      & 384 \\ \hline
27    & pw       & 256 & 1           & 1      & 768 \\ \hline
28    & dw       & $\infty$ & 5           & 1      & 768 \\ \hline
29    & pw       & 256 & 1           & 1      & 384 \\ \hline
30    & upsample & -   & -           & -      & -   \\ \hline
31    & dw       & $\infty$ & 5           & 1      & 320 \\ \hline
32    & pw       & 128 & 1           & 1      & 192 \\ \hline
33    & upsample & -   & -           & -      & -   \\ \hline
34    & dw       & $\infty$ & 5           & 1      & 160 \\ \hline
35    & pw       & 64  & 1           & 1      & 112 \\ \hline
36    & upsample & -   & -           & -      & -   \\ \hline
37    & dw       & $\infty$ & 3           & 1      & 112 \\ \hline
38    & pw       & 32  & 1           & 1      & 48  \\ \hline
39    & upsample & -   & -           & -      & -   \\ \hline
40    & dw       & $\infty$ & 3           & 1      & 28  \\ \hline
41    & pw       & 16  & 1           & 1      & 24  \\ \hline
42    & upsample & -   & -           & -      & -   \\ \hline
43    & pw       & 0   & 1           & 1      & 1   \\ \toprule
\end{tabular}
\caption{The discovered 87ms DNN architecture of NetAdaptV2 on NYU Depth V2 presented in Table~\ref{tab:fastdepth}. Type: type of the layer, which can be standard convolution (conv), depthwise convolution (dw), pointwise convolution (pw), or nearest neighbor upsampling (upsample). Filter: number of filters. All layers except for upsampling layers are followed by a batch normalization layer and a ReLU activation layer.}
\label{tab:discovered_architecture_fastdepth}
\end{table*}

\section{Formulation of Channel-Level Bypass Connections}

The formulation of channel-level bypass connections (CBCs), $Z = max(min(C, T), M)$, can be derived by considering the case 1 to 3 in Sec.~\ref{subsec:channel_level_bypass_connections} and Fig.~\ref{fig:cbc}. For the case 1 ($C = T$) and 2 ($C < T$), CBCs start bypassing input channels when $M$ becomes smaller than $C$ ($M < C$) to maintain the number of output channels $Z = max(C, M) = C$. For the case 3 ($C > T$), CBCs start bypassing input channels when $M$ becomes smaller than $T$ ($M < T$) instead of $C$, which requires replacing the $C$ in $Z = max(C, M)$ with $min(C, T)$ and gives the formulation of CBCs.

\section{Ablation Study on MobileNetV1}

This section provides the ablation study on MobileNetV1~\cite{Howard2017MobileNetV1}. This ablation study employs the experiment setup outlined in Sec.~\ref{subsec:experiment_setup} unless otherwise stated. The initial network is the largest MobileNetV1 (1.0 MobileNet-224~\cite{Howard2017MobileNetV1}).

\subsection{Impact of Channel-Level Bypass Connections}
The proposed channel-level bypass connections (CBCs) enable NetAdaptV2 to search for different network depths with marginal overhead. Table~\ref{tab:ablation_cbc_optimizer_mnv1} shows that supporting CBCs only increases the training time of the super-network by $1.2\times$. Moreover, the ability to search network depth allows discovering DNNs with better performance. As shown in Table~\ref{tab:ablation_cbc_optimizer_mnv1}, CBCs improve the accuracy of the discovered DNN by $6.5\%$ with the same latency.

\begin{table*}[t]
\centering
\begin{tabular}{c|c||c|c|c|c}
\toprule
\multicolumn{2}{c||}{Methods} & \multirow{2}{*}{\begin{tabular}[c]{@{}c@{}}Top-1\\ Accuracy (\%)\end{tabular}} & \multirow{2}{*}{\begin{tabular}[c]{@{}c@{}}\# of\\ Layers\end{tabular}} & \multirow{2}{*}{\begin{tabular}[c]{@{}c@{}}Super-Network Training \\ Speed (min/epoch)\end{tabular}} & \multirow{2}{*}{\# of Samples} \\ \cline{1-2}
CBC           & MCD           &                                                                          &                                                                          &                                                                                          &                                                                           \\ \toprule
      &              & 40.0 (+0)                                                                                                                                                & 28 (-0)                                                                      & 3.2 (100\%)                                                                                      & 1064 (100\%)                                                                      \\ \hline
      \checkmark       &              & 46.5 (+6.5)                                                                                                                                                & 19 (-9)                                                                       & 3.8 (119\%)                                                                                      & 1092 (103\%)                                                                      \\ \hline
      \checkmark       &       \checkmark       & 49.3 (+9.3)                                                                                                                                                 & 17 (-11)                                                                      & 3.8 (119\%)                                                                                      & 567 (53\%)                                                                       \\ \toprule
\end{tabular}
\caption{The ablation study of the channel-level bypass connections (CBCs) and the multi-layer coordinate descent (MCD) optimizer on ImageNet and MobileNetV1. The latency of the discovered networks is around 6.5ms.}
\label{tab:ablation_cbc_optimizer_mnv1}
\end{table*}

\subsection{Impact of Multi-Layer Coordinate Descent Optimizer}

The proposed multi-layer coordinate descent (MCD) optimizer improves the performance of the discovered DNN while reducing the number of samples and hence the search time. In this experiment, the MCD optimizer generates 27 samples ($J=27$) in each iteration, where $J$ is equal to the number of layers, and each sample is obtained by randomly shrinking 4 layers ($L=4$). Table~\ref{tab:ablation_cbc_optimizer_mnv1} shows that the MCD optimizer reduces the time for evaluating samples by $1.9\times$ and improves the accuracy by 2.8\%.

\section{Estimation of $CO_2$ Emission}
We estimate $CO_2$ emission based on Strubell et al.~\cite{strubell_2019_energy}. According to Table 3 in this paper, when BERT$_{base}$ is trained on 64 V100 GPUs for 79 hours, the $CO_2$ emission is 1438 lbs. Therefore, the ratio of $CO_2$ emission to GPU-hours is $\frac{1438}{64 \times 79} = 0.2844$. For each NAS method, we multiply its search time by this ratio to estimate its corresponding $CO_2$ emission.

\clearpage
\clearpage
{
\bibliographystyle{ieeetr}

\begin{thebibliography}{10}

\bibitem{eccv2018-netadapt}
T.-J. Yang, A.~Howard, B.~Chen, X.~Zhang, A.~Go, M.~Sandler, V.~Sze, and
  H.~Adam, ``{NetAdapt: Platform-Aware Neural Network Adaptation for Mobile
  Applications},'' in {\em European Conference on Computer Vision (ECCV)},
  2018.

\bibitem{Tan2018MnasNetPN}
M.~Tan, B.~Chen, R.~Pang, V.~Vasudevan, and Q.~V. Le, ``Mnasnet: Platform-aware
  neural architecture search for mobile,'' in {\em IEEE Conference on Computer
  Vision and Pattern Recognition (CVPR)}, 2019.

\bibitem{cai2018proxylessnas}
H.~Cai, L.~Zhu, and S.~Han, ``Proxyless{NAS}: Direct neural architecture search
  on target task and hardware,'' in {\em International Conference on Learning
  Representations (ICLR)}, 2019.

\bibitem{Chen2020MnasFPNLL}
B.~Chen, G.~Ghiasi, H.~Liu, T.-Y. Lin, D.~Kalenichenko, H.~Adam, and Q.~V. Le,
  ``Mnasfpn: Learning latency-aware pyramid architecture for object detection
  on mobile devices,'' {\em IEEE Conference on Computer Vision and Pattern
  Recognition (CVPR)}, pp.~13604--13613, 2020.

\bibitem{chamnet}
X.~{Dai}, P.~{Zhang}, B.~{Wu}, H.~{Yin}, F.~{Sun}, Y.~{Wang}, M.~{Dukhan},
  Y.~{Hu}, Y.~{Wu}, Y.~{Jia}, P.~{Vajda}, M.~{Uyttendaele}, and N.~K. {Jha},
  ``Chamnet: Towards efficient network design through platform-aware model
  adaptation,'' in {\em IEEE Conference on Computer Vision and Pattern
  Recognition (CVPR)}, pp.~11390--11399, 2019.

\bibitem{zoph2017nasreinforcement}
B.~Zoph and Q.~V. Le, ``Neural architecture search with reinforcement
  learning,'' in {\em International Conference on Learning Representations
  (ICLR)}, 2017.

\bibitem{zoph2018nasnet}
B.~Zoph, V.~Vasudevan, J.~Shlens, and Q.~V. Le, ``Learning transferable
  architectures for scalable image recognition,'' {\em IEEE Conference on
  Computer Vision and Pattern Recognition (CVPR)}, 2018.

\bibitem{yu2018slimmable}
J.~Yu, L.~Yang, N.~Xu, J.~Yang, and T.~Huang, ``Slimmable neural networks,'' in
  {\em International Conference on Learning Representations (ICLR)}, 2019.

\bibitem{Yu_2019_ICCV}
J.~Yu and T.~S. Huang, ``Universally slimmable networks and improved training
  techniques,'' in {\em International Conference on Computer Vision (ICCV)},
  October 2019.

\bibitem{autoslim_arxiv}
J.~Yu and T.~Huang, ``Autoslim: Towards one-shot architecture search for
  channel numbers,'' {\em ArXiv}, vol.~abs/1903.11728, 2019.

\bibitem{cai2020once}
H.~Cai, C.~Gan, T.~Wang, Z.~Zhang, and S.~Han, ``Once for all: Train one
  network and specialize it for efficient deployment,'' in {\em International
  Conference on Learning Representations (ICLR)}, 2020.

\bibitem{yu2020bignas}
J.~Yu, P.~Jin, H.~Liu, G.~Bender, P.-J. Kindermans, M.~Tan, T.~Huang, X.~Song,
  R.~Pang, and Q.~Le, ``Bignas: Scaling up neural architecture search with big
  single-stage models,'' in {\em European Conference on Computer Vision
  (ECCV)}, 2020.

\bibitem{Bender2018UnderstandingAS}
G.~Bender, P.~Kindermans, B.~Zoph, V.~Vasudevan, and Q.~V. Le, ``Understanding
  and simplifying one-shot architecture search,'' in {\em International
  Conference on Machine Learning (ICML)}, 2018.

\bibitem{enas}
H.~Pham, M.~Y. Guan, B.~Zoph, Q.~V. Le, and J.~Dean, ``Efficient neural
  architecture search via parameter sharing,'' in {\em International Conference
  on Machine Learning (ICML)}, 2018.

\bibitem{tunas}
G.~Bender, H.~Liu, B.~Chen, G.~Chu, S.~Cheng, P.-J. Kindermans, and Q.~V. Le,
  ``Can weight sharing outperform random architecture search? an investigation
  with tunas,'' in {\em IEEE Conference on Computer Vision and Pattern
  Recognition (CVPR)}, June 2020.

\bibitem{Guo2020SPOS}
Z.~Guo, X.~Zhang, H.~Mu, W.~Heng, Z.~Liu, Y.~Wei, and J.~Sun, ``Single path
  one-shot neural architecture search with uniform sampling,'' in {\em European
  Conference on Computer Vision (ECCV)}, 2020.

\bibitem{gordon2018morphnet}
A.~Gordon, E.~Eban, O.~Nachum, B.~Chen, T.-J. Yang, and E.~Choi, ``Morphnet:
  Fast \& simple resource-constrained structure learning of deep networks,'' in
  {\em IEEE Conference on Computer Vision and Pattern Recognition (CVPR)},
  2018.

\bibitem{liu2018darts}
H.~Liu, K.~Simonyan, and Y.~Yang, ``{DARTS}: Differentiable architecture
  search,'' in {\em International Conference on Learning Representations
  (ICLR)}, 2019.

\bibitem{wu2018fbnet}
B.~Wu, X.~Dai, P.~Zhang, Y.~Wang, F.~Sun, Y.~Wu, Y.~Tian, P.~Vajda, Y.~Jia, and
  K.~Keutzer, ``Fbnet: Hardware-aware efficient convnet design via
  differentiable neural architecture search,'' {\em IEEE Conference on Computer
  Vision and Pattern Recognition (CVPR)}, 2019.

\bibitem{fbnetv2}
A.~{Wan}, X.~{Dai}, P.~{Zhang}, Z.~{He}, Y.~{Tian}, S.~{Xie}, B.~{Wu}, M.~{Yu},
  T.~{Xu}, K.~{Chen}, P.~{Vajda}, and J.~E. {Gonzalez}, ``Fbnetv2:
  Differentiable neural architecture search for spatial and channel
  dimensions,'' in {\em IEEE Conference on Computer Vision and Pattern
  Recognition (CVPR)}, pp.~12962--12971, 2020.

\bibitem{stamoulis2019singlepath}
D.~Stamoulis, R.~Ding, D.~Wang, D.~Lymberopoulos, B.~Priyantha, J.~Liu, and
  D.~Marculescu, ``Single-path nas: Designing hardware-efficient convnets in
  less than 4 hours,'' in {\em arXiv preprint arXiv:1904.02877}, 2019.

\bibitem{stamoulis2019singlepathautoml}
D.~Stamoulis, R.~Ding, D.~Wang, D.~Lymberopoulos, B.~Priyantha, J.~Liu, and
  D.~Marculescu, ``Single-path mobile automl: Efficient convnet design and nas
  hyperparameter optimization,'' in {\em arXiv preprint arXiv:1907.00959},
  2019.

\bibitem{Mei2020AtomNAS}
J.~Mei, Y.~Li, X.~Lian, X.~Jin, L.~Yang, A.~Yuille, and J.~Yang, ``Atomnas:
  Fine-grained end-to-end neural architecture search,'' in {\em International
  Conference on Learning Representations (ICLR)}, 2020.

\bibitem{Xu2020PC-DARTS}
Y.~Xu, L.~Xie, X.~Zhang, X.~Chen, G.-J. Qi, Q.~Tian, and H.~Xiong, ``Pc-darts:
  Partial channel connections for memory-efficient architecture search,'' in
  {\em International Conference on Learning Representations (ICLR)}, 2020.

\bibitem{Howard_2019_ICCV}
A.~Howard, M.~Sandler, G.~Chu, L.-C. Chen, B.~Chen, M.~Tan, W.~Wang, Y.~Zhu,
  R.~Pang, V.~Vasudevan, Q.~V. Le, and H.~Adam, ``Searching for mobilenetv3,''
  in {\em International Conference on Computer Vision (ICCV)}, 2019.

\bibitem{book2020sze}
V.~Sze, Y.-H. Chen, T.-J. Yang, and J.~Emer, {\em {Efficient Processing of Deep
  Neural Networks}}.
\newblock {Morgan \& Claypool}, 2020.

\bibitem{imagenet_cvpr09}
J.~Deng, W.~Dong, R.~Socher, L.-J. Li, K.~Li, and L.~Fei-Fei, ``{ImageNet: A
  Large-Scale Hierarchical Image Database},'' in {\em IEEE Conference on
  Computer Vision and Pattern Recognition (CVPR)}, 2009.

\bibitem{nyudepth}
P.~K. Nathan~Silberman, Derek~Hoiem and R.~Fergus, ``Indoor segmentation and
  support inference from rgbd images,'' in {\em European Conference on Computer
  Vision (ECCV)}, 2012.

\bibitem{Sandler_2018_CVPR}
M.~Sandler, A.~Howard, M.~Zhu, A.~Zhmoginov, and L.-C. Chen, ``Mobilenetv2:
  Inverted residuals and linear bottlenecks,'' in {\em IEEE Conference on
  Computer Vision and Pattern Recognition (CVPR)}, 2018.

\bibitem{dropout}
N.~Srivastava, G.~Hinton, A.~Krizhevsky, I.~Sutskever, and R.~Salakhutdinov,
  ``Dropout: A simple way to prevent neural networks from overfitting,'' {\em
  Journal of Machine Learning Research}, vol.~15, no.~56, pp.~1929--1958, 2014.

\bibitem{Tan2019EfficientNet}
M.~Tan and Q.~V. Le, ``Efficientnet: Rethinking model scaling for convolutional
  neural networks,'' in {\em International Conference on Machine Learning
  (ICML)}, 2019.

\bibitem{ijcv2015-russakovsky-ilsvrc}
O.~Russakovsky, J.~Deng, H.~Su, J.~Krause, S.~Satheesh, S.~Ma, Z.~Huang,
  A.~Karpathy, A.~Khosla, M.~Bernstein, A.~C. Berg, and L.~Fei-Fei, ``{ImageNet
  Large Scale Visual Recognition Challenge},'' {\em International Journal of
  Computer Vision (IJCV)}, vol.~115, no.~3, pp.~211--252, 2015.

\bibitem{Goyal2017AccurateLM}
P.~Goyal, P.~Doll{\'a}r, R.~B. Girshick, P.~Noordhuis, L.~Wesolowski,
  A.~Kyrola, A.~Tulloch, Y.~Jia, and K.~He, ``Accurate, large minibatch sgd:
  Training imagenet in 1 hour,'' {\em arXiv}, vol.~abs/1706.02677, 2017.

\bibitem{strubell_2019_energy}
E.~Strubell, A.~Ganesh, and A.~McCallum, ``Energy and policy considerations for
  deep learning in {NLP},'' in {\em Proceedings of the 57th Annual Meeting of
  the Association for Computational Linguistics}, pp.~3645--3650, 2019.

\bibitem{chu2019fairnas}
X.~Chu, B.~Zhang, and R.~Xu, ``Fairnas: Rethinking evaluation fairness of
  weight sharing neural architecture search,'' {\em arXiv preprint
  arXiv:1907.01845}, 2019.

\bibitem{autoaugment}
E.~D. Cubuk, B.~Zoph, D.~Mane, V.~Vasudevan, and Q.~V. Le, ``Autoaugment:
  Learning augmentation strategies from data,'' in {\em IEEE Conference on
  Computer Vision and Pattern Recognition (CVPR)}, June 2019.

\bibitem{stochastic_depth}
G.~Huang, Y.~Sun, Z.~Liu, D.~Sedra, and K.~Q. Weinberger, ``Deep networks with
  stochastic depth,'' in {\em European Conference on Computer Vision (ECCV)},
  2016.

\bibitem{lu2020nsganetv2}
Z.~Lu, K.~Deb, E.~Goodman, W.~Banzhaf, and V.~N. Boddeti, ``{NSGANetV2}:
  Evolutionary multi-objective surrogate-assisted neural architecture search,''
  in {\em European Conference on Computer Vision (ECCV)}, 2020.

\bibitem{Tan2019MixConvMD}
M.~Tan and Q.~V. Le, ``Mixconv: Mixed depthwise convolutional kernels,'' in
  {\em British Machine Vision Conference (BMVC)}, 2019.

\bibitem{Howard2017MobileNetV1}
A.~G. Howard, M.~Zhu, B.~Chen, D.~Kalenichenko, W.~Wang, T.~Weyand,
  M.~Andreetto, and H.~Adam, ``Mobilenets: Efficient convolutional neural
  networks for mobile vision applications,'' {\em ArXiv}, vol.~abs/1704.04861,
  2017.

\bibitem{li2020reproducibility}
L.~Li and A.~Talwalkar, ``Random search and reproducibility for neural
  architecture search,'' in {\em Proceedings of Machine Learning Research},
  vol.~115, pp.~367--377, 22--25 Jul 2020.

\bibitem{icra_2019_fastdepth}
{Wofk, Diana and Ma, Fangchang and Yang, Tien-Ju and Karaman, Sertac and Sze,
  Vivienne}, ``{FastDepth: Fast Monocular Depth Estimation on Embedded
  Systems},'' in {\em International Conference on Robotics and Automation
  (ICRA)}, {2019}.

\end{thebibliography}

}
\end{document}